\documentclass[11pt]{article}
\usepackage[margin=1in]{geometry}
\pdfoutput=1
\usepackage{url}

\usepackage[utf8]{inputenc} 
\usepackage[T1]{fontenc}    
\usepackage[colorlinks=true,linkcolor=blue,citecolor=blue]{hyperref}   
\usepackage{url}            
\usepackage{booktabs}       
\usepackage{amsfonts}       
\usepackage{nicefrac}       
\usepackage{microtype}      
\usepackage{amsmath,amssymb, amsthm, mathtools, dsfont}
\usepackage{cleveref}       
\usepackage{graphicx}

\usepackage{enumitem}
\usepackage{comment}
\usepackage{bbm}
\usepackage{natbib}

\usepackage{definitions}
\usepackage{color}
\usepackage{wrapfig}
\usepackage{multirow}
\usepackage{tabularx}
\usepackage{algorithm}
\usepackage{algpseudocode}
\usepackage{multirow}
\usepackage{tabularx}
\usepackage{array}
\usepackage{natbib}
\usepackage{mathabx}
\usepackage{graphicx}
\usepackage{rotating}
\usepackage{multirow}

\usepackage{amsfonts}
\usepackage{caption}
\usepackage{subcaption}
\usepackage{amsmath}
\usepackage{amssymb}
\usepackage{pifont}
\usepackage{bbm}
\usepackage{dsfont}
\usepackage[usenames,dvipsnames,svgnames,table]{xcolor}

\numberwithin{equation}{section}
\numberwithin{theorem}{section}

\usepackage{xargs}

\usepackage{booktabs} 
\usepackage{multirow}
\usepackage{float}
\usepackage{placeins}
\usepackage{tikz}
\usetikzlibrary{bayesnet}
\usetikzlibrary{arrows}
\usepackage{color}
\usepackage{graphicx}
\usepackage{caption}
\usepackage{subcaption}
\usetikzlibrary{backgrounds}
\usetikzlibrary{bayesnet}
\usetikzlibrary{arrows}
\usepackage{standalone}
\usepackage{pdflscape}
\usepackage{bibunits}
\DeclareMathOperator{\VEC}{vec}
\newcommand{\bo}{\mathbf{o}}
\newcommand{\kt}[1]{\textcolor{ForestGreen}{Katherine: #1}}

\title{Proxy Methods for Domain Adaptation}

\author{ Katherine Tsai$^1$, Stephen R. Pfohl$^2$, Olawale Salaudeen$^1$,\\
Nicole Chiou$^3$, Matt J. Kusner$^4$,\\
Alexander D'Amour$^5$, Sanmi Koyejo$^{3,5}$, Arthur Gretton$^{5,6}$ }

\date{
$^1$University of Illinois Urbana-Champaign \\
$^2$ Google Research\\
$^3$Stanford University \\
$^4$University College London \\
$^5$Google DeepMind \\
$^6$Gatsby Computational Neuroscience Unit 
}

\begin{document}

\begin{bibunit}[my-plainnat]
\maketitle
\begin{abstract}

We study the problem of domain adaptation under distribution shift, where the shift is due to a change in the distribution of an unobserved, latent variable that confounds both the covariates and the labels. 
In this setting, neither the covariate shift nor the label shift assumptions apply. 
Our approach to adaptation employs proximal causal learning, a technique for estimating causal effects in settings where proxies of unobserved confounders are available.
We demonstrate that proxy variables allow for adaptation to distribution shift without explicitly recovering or modeling latent variables. 
We consider two settings, (i) 
\textbf{Concept Bottleneck}: an additional ``concept'' variable is observed that mediates the relationship between the covariates and labels;
(ii) \textbf{Multi-domain}: training data from multiple source domains is available, where each source domain exhibits a different distribution over the latent confounder.
We develop a two-stage kernel estimation approach to adapt to complex distribution shifts in both settings. In our experiments, we show that our approach outperforms other methods, notably those which explicitly recover the latent confounder.

\end{abstract}
\vspace{-0.2cm}
\section{Introduction}
\vspace{-0.2cm}

The goal of domain adaptation is to transfer an accurate model from a labeled \emph{source} domain to an unlabeled \emph{target} domain, which has a different but related distribution~\citep{pan2010domain,koh2021wilds, malinin2021shifts}. It is motivated by the fact that labeling data is often labor intensive, and sometimes requires domain expertise. For example, the distribution of patients diagnosed with a condition from hospital $A$ and hospital $B$ may differ due to patients’ socioeconomic status, demographics, and other factors. 
However, labeled data might be only be available at hospital $A$ and not at hospital $B$ (e.g., due to less funding).
As a result, an accurate model for patients from hospital $A$ may perform poorly for patients from hospital $B$.

In order to provide guarantees on the accuracy of a transferred model, one of two classical assumptions have been made: \emph{label shift} or \emph{covariate shift}. Label shift~\citep{buck1966comparison, lipton2018detecting} assumes that the distribution of a label $P(Y)$ shifts between source and target domains, but the conditional distribution $P(X\mid Y)$ does not. Conversely, covariate shift~\citep{shimodaira2000improving} assumes that the covariate distribution $P(X)$ shifts between domains, but the distribution $P(Y\mid X)$ stays the same. Each assumption provides theoretical guarantees on the generalization of a transferred classifier. In fact, without any assumptions, the source and target domains could differ arbitrarily, making guarantees impossible. 
However, these assumptions are often too restrictive to apply in real-world settings~\citep{zhang2015multi, schrouff2022diagnosing}.
For instance, if covariates $X$ and labels $Y$ are confounded by a third variable $U$, it is possible for neither $P(X\mid Y)$ or $P(Y\mid X)$ to be equal across domains. For example, demographic information $U$ could confound the relationship between a diagnosis $Y$ and a radiological image $X$. 
In this example, if two hospitals have different distributions over demographics, both label shift and covariate shift adaptation methods will fail to transfer a classifier across hospitals. 


To address this, recent work has introduced a \emph{latent shift} assumption: the distribution of $U$, an unobserved latent confounder of $X$ and $Y$, shifts between the source and target domain \citep{alabdulmohsin2023adapting}. 
In this setting, all distributions of $X$ and $Y$ (without conditioning on $U$) may differ across the domains, violating label and covariate shift assumptions. 

\textbf{Contributions.} 
We propose techniques for domain adaptation under the latent shift assumption that are guaranteed to identify the optimal predictor $\EE[Y\mid x]$ in the target domain. We make use of proxy methods \citep{miao2018identifying}, which are a recently developed framework for causal effect estimation in the presence of a hidden confounder $U$, given indirect proxy information on $U$. Compared to prior work \citep{alabdulmohsin2023adapting}, our techniques do not require: identifying the distribution of the latent variable $U$, that $U$ be discrete, 
or further linear independence assumptions. 
We consider two settings: (1) \textbf{Concept Bottleneck}: we observe in both domains a \textit{proxy} $W$ of the unobserved confounder $U$ and a \textit{concept} $C$ that mediates the direct relationship between $X$ and $Y$~\citep{alabdulmohsin2023adapting}, or (2) \textbf{Multi-Domain}: we do not observe $C$ in either domain, but have access to observations from multiple source domains.
For both settings, we provide  guarantees for identifying $\EE[Y\mid x]$ without observing $Y$ in the target domain. 
When $\EE[Y\mid x]$ is identifiable, we develop practical two-stage kernel estimators to perform adaptation.

\vspace{-0.2cm}
\section{Related Work}
\vspace{-0.2cm}
The development of techniques for learning robust models and adapting to distribution shift has a long history in machine learning, but recently has received increased attention \citep{shen2021towards,zhou2022domain,wang2022generalizing}. 

\textbf{Causality for domain adaptation.}
Our work is inspired by techniques that formulate the covariate/label shift settings as assumptions on the causal structure for domain adaptation and distributional robustness (\textit{e.g},  \citet{scholkopf2012causal,peters2015causal,zhang2015multi,subbaswamy2019preventing,rothenhausler2021anchor,veitch2021counterfactual,magliacane2018domain,arjovsky2019invariant,ganin2016domain,ben2010theory, oberst2021regularizing}).

\textbf{Proximal causal inference.}
Our identification technique is inspired by  approaches used to identify causal effects with unobserved confounding with observed proxies~\citep{kuroki2014measurement,miao2018identifying,deaner2018proxy,tchetgen2020introduction,mastouri2021proximal,cui2023semiparametric,xu2023kernel}. These approaches design `bridge functions' to connect quantities involving a proxy $W$ with those of the label $Y$. The beauty of this approach is that these bridge functions are implicitly a marginalization over $U$. This allows these approaches to identify causal quantities without identifying  distributions involving $U$.

\textbf{Latent shift.}
Our work is most closely related to \citet{alabdulmohsin2023adapting}, who introduced the setting of latent shift with proxies $W$ and concepts $C$. They showed that the optimal predictor $\EE[Y\mid x]$ is identifiable in the target domain if $W$ and $C$ are observed in the source domain and $X$ is observed in the target domain. To do so, they required (a) identification of distributions involving $U$, (b) that $U$ is a discrete variable, (c) knowledge of the dimensionality of $U$, and (d) additional linear independence assumptions. In contrast, our work derives identification results for arbitrary $U$, and does not require any of (a)-(d). However, there is no free lunch: to achieve this, we require that proxies $W$ are observed in the target, and either that: (i) concepts $C$ are also observed in the target, or (ii) we observe multiple source domains. For (ii) we do not require $C$ in either the source or the target, but for full identification we require that $U$ is discrete.

\vspace{-0.2cm}
\section{Problem Framework}\label{section:setup}
\vspace{-0.2cm}
\begin{figure*}
    \centering
    \small
    \begin{subfigure}[t]{0.19\textwidth}
    \resizebox{!}{7em}{\begin{tikzpicture}
     \node[latent, xshift=2.5cm, line width=.8pt] (Y) {$Y$};%
     \node[latent, line width=.8pt] (X) {$X$};%
     \node[obs, above=of X,xshift=1cm, line width=.8pt] (U) {$U$}; %
     \edge [line width=.8pt] {U} {X}
     \edge [line width=.8pt]{X} {Y}
     \end{tikzpicture}}
    \caption{Covariate shift}
    \label{fig:covshift}
    \end{subfigure}
    \begin{subfigure}[t]{0.19\textwidth}
    \resizebox{!}{7em}{\begin{tikzpicture}{
     \node[latent, xshift=2.5cm, line width=.8pt] (Y) {$Y$};%
     \node[latent, line width=.8pt] (X) {$X$};%
     \node[obs, above=of X,xshift=1cm, line width=.8pt] (U) {$U$}; %
     \edge [line width=.8pt] {U} {Y}
     \edge [line width=.8pt]{Y} {X}}
    \end{tikzpicture}}
    \caption{Label shift}
    \label{fig:labelshift}
    \end{subfigure}
    \begin{subfigure}[t]{0.28\textwidth}
    \resizebox{!}{7em}{\begin{tikzpicture}  
     \node[latent, line width=.8pt] (C) {$C$};%
     \node[latent, xshift=2cm, line width=.8pt] (Y) {$Y$};%
     \node[latent, above=of C,xshift=-1cm, line width=.8pt] (X) {$X$};%
     \node[obs, above=of C,xshift=1cm, line width=.8pt] (U) {$U$}; %
     \node[latent,above=of Y, xshift=1cm, line width=.8pt] (W) {$W$} edge [->,line width=.8pt] (Y);%
     \edge [line width=.8pt] {U} {X}
     \edge [line width=.8pt]{U} {W}
     \edge [line width=.8pt]{U, X} {C}
     \edge [line width=.8pt]{C, U} {Y}
    \end{tikzpicture}}

    \caption{Concept Bottleneck shift}
    \label{fig:shiftwconcepts}
    \end{subfigure}
    \begin{subfigure}[t]{0.28\textwidth}
    \resizebox{!}{7 em}{\begin{tikzpicture}
     \node[latent, line width=.8pt] (X) {$X$};%
     \node[latent, xshift=2cm, line width=.8pt] (Y) {$Y$};%
     \node[obs, above=of C,xshift=-1cm, line width=.8pt] (Z) {$Z$};%
     \node[obs, above=of C,xshift=1cm, line width=.8pt] (U) {$U$}; %
     \node[latent,above=of Y, xshift=1cm, line width=.8pt] (W) {$W$} edge [->, line width=.8pt] (Y);%
     \edge [line width=.8pt] {U} {X}
     \edge [line width=.8pt]{U} {W}
     \edge [line width=.8pt]{X, U} {Y}
     \edge [line width=.8pt] {Z} {U}
    \node [fit=(X) (Y) (U) (W), inner sep=3pt, draw, line width=.5pt, rounded corners] (plate1) {};
    \node [above left, inner sep=1pt, xshift=-4pt, yshift=2pt] at (plate1.south east) {\large $n$};
     \end{tikzpicture}}
     \caption{Multi-Domain shift}
     \label{fig:multishift}
     \end{subfigure}
     
\caption{Causal diagrams. The shaded circle denotes unobserved variable and the solid circle denotes observed variable. $X$ is the covariate, $Y$ is the response, $C$ is the concept, $W$ is the proxy, $Z$ is the domain-related variable, and $U$ is the latent variable.  
}
\label{fig:model}
\end{figure*}
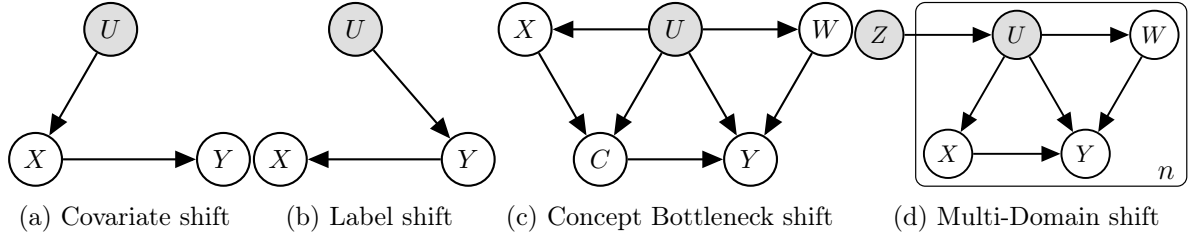


Let $P(\cdot)$ and $Q(\cdot)$ denote the probability distribution functions of the source domain and target domain, respectively. Let $p$ and $q$ indicate source and target quantities. Our goal is to study identification and estimation of the optimal target predictor $\EE_q[Y\mid x]$ when $Y$ is not observed in the target domain.

\textbf{Concept Bottleneck.} The first setting we study is described by the graph in Figure~\ref{fig:shiftwconcepts}. 
We have two additional variables: (i) proxies $W$, which provide auxiliary information about $U$, or can be seen as a noisy version of it~\citep{kuroki2014measurement}, and (ii) concepts $C$, which mediate or `bottleneck' the relationship between the covariates $X$ and labels $Y$~\citep{goyal2019explaining,koh2020concept}. 
For example, \citet{koh2020concept} describe a setting where the concepts $C$ are high-level clinical and morphological features of a knee X-ray $X$, which mediate the relationship with osteoporosis severity $Y$. In this example, $U$ could describe demographic variations that alter symptoms $X,C$ and outcome $Y$, and the proxies $W$ could include patient background and clinical history (e.g., prior diagnoses, medications, procedures, etc). For the source domain we assume we observe $(X,C,W,Y) \!\sim\! P$ and for the target domain we observe $(X,C,W) \!\sim\! Q$.
We formalize the notion of latent shift, as introduced in \citet{alabdulmohsin2023adapting}. 

\begin{assumption}[Concept Bottleneck, \cite{alabdulmohsin2023adapting}]\label{assumption:graph}
The shift between $P$ and $Q$ is located in unobserved $U$, i.e., there is a latent shift $P(U)\not\eq Q(U)$, but $P(V \mid U)=Q(V\mid U)$, where $V\subseteq\{W,X,C,Y\}$. 
\end{assumption}

This assumption states that every variable conditioned on $U$ is invariant across domains. However, as $P(U)\!\not\eq\! Q(U)$, none of the marginal distributions are: $P(V)\!\not\eq\! Q(V)$ for $V\!\subseteq\!\{W,X,C,Y\}$. This assumption is a generalization of  covariate shift $P(Y\mid X,U)\!=\!Q(Y\mid X,U)$~\citep{shimodaira2000improving} and  label shift $P(X\mid Y, U)\!=\!Q(X\mid Y, U)$~\citep{buck1966comparison}, with associated graphs in Figure~\ref{fig:covshift}--\ref{fig:labelshift}.

\begin{assumption}[Structural assumption]\label{assumption:CI}
Graphs in Figure~\ref{fig:model} are faithful and Markov~\citep{spirtes2000causation}.
\end{assumption}
Under Assumption~\ref{assumption:CI}, we have the following conditional independence properties for the graph in Figure~\ref{fig:shiftwconcepts}:
\[
 Y\indep X\mid\{U,C\},\quad W\indep\{X,C\}\mid U.
\]
With this conditional independence structure, $\{U, C\}$ blocks the information from $X$ to $Y$ and $U$ blocks the information flow from $W$ to $\{X, C\}$.
We will see in Section \ref{ssec:identification} that these assumptions allow us to obtain $Q(Y\mid x)$ from $Q(W,C\mid x)$ in the target domain, where the latter is a function of observed quantities.


\textbf{Multi-domain}. In the second setting, suppose we do not observe the concepts $C$ in any domain, but instead observe data from multiple source domains, according to the graph in Figure~\ref{fig:multishift}. 
For instance, we may want to learn a classifier for a target hospital that has only unlabelled data, using data from several source hospitals with labelled data.
Here, let $Z$ be a random variable in $\Zcal$ denoting a prior over the source domains, and let $P(U\vert Z)$ be the distribution of $U$ given $Z$. We make $k_Z$ draws from $Z$, indexed by $r\in \{1,\ldots, k_Z\}$, and write $\{z_1, \ldots, z_{k_Z}\}=:\Zcal_p \subseteq \Zcal$. 
For each source domain $z_r$, 
we observe $(X,W,Y) \!\sim\! P(X,W,Y\vert z_r) \!:=\! P_r(X,W,Y)$. For the target, we denote it with index $k_{Z}+1$ and only observe $(X,W) \!\sim\! P(X,W \vert z_{k_Z+1}) \!:=\! Q(X,W)$. In general let $P_r(V)\!:=\!P(V|z_r)$ and $Q(V)\!:=\!P(V|z_{k_Z+1})$ for any $V\subseteq\{W,X,Y,U\}$. 
For this setting we replace Assumption~\ref{assumption:graph} with the following shift assumption.

\begin{assumption}[Multi-Domain]\label{assumption:uz_direction} For each $z,z'\in\Zcal_p$ such that $z\not\eq z'$, we have $P(U\vert z)\not\eq P(U\vert z')\neq Q(U)$. 
\end{assumption}
Note that Assumption~\ref{assumption:CI} implies the following the conditional independence property in Figure~\ref{fig:multishift}:
\[
\{Y,X,W\}\indep Z\mid U.
\]
Under Assumption~\ref{assumption:uz_direction}, we allow all joint distributions to be different $$P(W,X,U,Y\vert z)\not\eq P(W,X,U,Y\vert z') \neq Q(W,X,U,Y)$$ for $z\neq z'\in\Zcal_p$.


\vspace{-0.2cm}
\section{Identification under Latent Shifts}\label{ssec:identification}
\vspace{-0.2cm}
Our identification techniques are inspired by proximal causal inference~\citep{tchetgen2020introduction}. The key idea is to design so-called ``bridge'' functions to identify distributions confounded by unobserved variables.  We first show that with additional proxies and concepts, $\EE_{q}[Y\mid x]$ is identifiable under any latent shift. 
\subsection{Identification with Concepts}\label{ssec:identificationwconcept}

To prove identifiability, we need certain assumptions to hold for the shift. The first is a regularity assumption, also known as a completeness condition, and is commonly used to identify causal estimands~\citep{d2011completeness, miao2018identifying}.

\begin{assumption}[Informative variables] \label{assumption:completeness} Let $g$ be any mean squared integrable function. Both the source domain and the target domain, $(f,F)\in\{(p,P), (q,Q)\}$, satisfy 
$
    \EE_f[g(U)\mid x,c] = 0 
$ for all $x\in\Xcal, c\in\Ccal$ if and only if $g(U)=0$ almost surely with respect to $F(U)$.
\end{assumption}
At a high level, completeness states that the $X$ must have sufficient variability related to the change of $U$. This is a common assumption made in proximal causal inference~(cf. Condition~(ii) in ~\citet{miao2018identifying} and Assumption~3 in~\citet{mastouri2021proximal}).
For more details on the justification of completeness assumption, see the supplementary material of~\citet{miao2022identifying}. 
 
Second, we need a guarantee on the support of $u\in\Ucal$. Intuitively, if a $u\in\Ucal$ has non-zero probability in the target domain, it should have non-zero probability in the source domain as well. Otherwise, it is impossible to adjust to certain shifts (as we never see these regimes in the source domain). This is similar to the positivity assumption commonly made in causality literature~\citep{hernan2006estimating}. 

 
 \begin{assumption}[Positivity]\label{assumption:subsetsupport}
For any $u\in\Ucal$, if $Q(u) > 0$ then $P(u) > 0$.
 \end{assumption}

If data are generated according to Figure~\ref{fig:shiftwconcepts}, and the regularity conditions~\ref{assumption:completeness_h0}--\ref{assumption:rc_h0_2} hold (see Appendix~\ref{appendix:existence:h_0}), \citet{miao2018identifying} first showed the existence of the solutions $h_0^p(w,c), h_0^q(w,c)$ of the following equations:
\begin{align}\label{eq:h0}
\EE_p[Y\mid c,x]=&\;\int_{\Wcal}h_0^p(w,c)\mathrm{d}P(w\mid c,x) \\
\EE_q[Y\mid c,x]=&\;\int_{\Wcal}h_0^q(w,c)\mathrm{d}Q(w\mid c,x). \nonumber
\end{align}
The terms $h_0^p(w,c), h_0^q(w,c)$ are called `bridge' functions as they connect the proxy $W$ to the label $Y$. 
If we are able to identify $h_0^q(w,c)$ then we can identify $\EE_q[Y \mid x]$, by using eq.~(\ref{eq:h0}) to obtain $\EE_q[Y \mid C, x]$ and marginalizing over $Q(C \mid x)$. 


We show that it is possible to connect identification of $h_0^q(w,c)$ with that of $h_0^p(w,c)$, leading directly to identification of $\EE_q[Y\mid x]$.
\begin{theorem}\label{theorem:complete_identification}
Assume that $h_0^p$ and $h_0^q$ exist (i.e., regularity Assumptions~\ref{assumption:completeness_h0}--\ref{assumption:rc_h0_2} hold). 
Then given Assumptions~\ref{assumption:graph}, \ref{assumption:CI}, \ref{assumption:completeness}, \ref{assumption:subsetsupport} we have that, for any $c\in\Ccal$,  $$\int_{\Wcal} h_0^p(w,c)\mathrm{d}P(w\mid u)=\int_{\Wcal} h_0^q(w,c)\mathrm{d}Q(w\mid u),
$$ almost surely with respect to $Q(U)$. This implies that 
$$\EE_q[Y\mid x]=\int_{\Wcal\times \Ccal} h_0^p(w,c)\mathrm{d}Q(w,c\mid x).$$
\end{theorem}
The proof is given in Appendix \ref{ssec:theorem:concept}. Hence, given $h_0^p$ and $(W,X,C)$ from the target $Q$, we are able to adapt to arbitrary distribution shifts in unobserved $U$. The advantage of this approach is that it will not require estimating any distributions involving $U$. We demonstrate this in Section~\ref{sec:kernelapproach}.

While concepts can ensure identifiability,
they may not be available in practice. In this case, a natural question is whether the optimal target predictor $\EE_q[Y\mid x]$ is still identifiable. In the next section we show that if we instead have access to data from multiple source domains, $\EE_q[Y\mid x]$ may again be identifiable. 

\subsection{The Blessings of Multiple Domains}\label{sec:blessingsmultipledomains}

We now turn to the multi-domain setting. 
The graphical structure in Figure~\ref{fig:multishift} is similar to the structure in Figure~\ref{fig:shiftwconcepts} with $C$ replaced by $X$, $X$ replaced by $Z$, and the arrow between $U$ and $Z$ flipped. Although the bridge function proposed by~\citet{miao2018identifying} assumes an edge from $U$ to $Z$, changing the direction from $Z$ to $U$ does not change the conditional independence structure~\citep{pearl2009causality}. The main difference is we will only be able to guarantee full identification when $U$ is discrete. We start by demonstrating this, and then give an example of the inherent difficulty of identification when $U$ is continuous.

To begin, for simplicity, assume $U$ and $W$ are discrete (with dimensionalities $k_U$ and $k_W$). We have finitely many samples from $Z$, denoted as $z_1,\ldots, z_{k_Z}$, corresponding to our training domains.  
We seek a bridge function (in this case, a matrix $M_0(w_i,x)$)  satisfying
\begin{align}
\EE_r[Y\mid x] = \sum_{i=1}^{k_w}M_0(w_i,x)P_r(w_i\mid x),\label{eq:bridgeM0}
\end{align}
for all $r=1,\ldots, k_Z$, where $\EE_r[Y\mid x]$ is the conditional expectation obtained in domain $r$, and $P_r(W\mid x)= P(W\mid x,z_r)$. 

In order to identify $M_0(w_i,x)$, and then $\EE_q[Y\mid x]$, we need enough source domains to capture the variability of $U$. The following result describes how many we need.


\begin{proposition}\label{prop:numenv}  Suppose that we have $k_Z$ source domains and $W$, $U$ have $k_W$ and $k_U$ categories respectively. Then, if $k_W, k_Z\geq k_U$ and subject to appropriate rank conditions (see proof in Appendix~\ref{ssec:prop:numenv}),  the bridge function is identifiable and does not depend on the specific $z$. 
\end{proposition}
This result generalizes the identification analysis developed in~\citet{miao2018identifying}. If the number of observed source domains $k_Z$ is greater than the dimension of the latent $U$, then subject to appropriate identifiability requirements (detailed in Appendix~\ref{ssec:prop:numenv}), we can recover the bridge  $M_0(w_i,x)$.

Now, consider the case where $U$ is discrete but all observed variables $W,X,Y$ are continuous. In this case we have the following system
\begin{equation}
\EE_r[Y\mid x] =\int_{\Wcal} m_0(w,x)\mathrm{d}P_r(w\mid x), \label{eq:multiDomainContinuous}
\end{equation}
for $r=1,\ldots, k_Z$. The proof of existence of $m_0$ is a modification of Proposition~\ref{appendix:existence:h_0}, as shown in Proposition~\ref{appendix:existence:m_0}. In order to identify target $\EE_q[Y\mid x]$, we need the following assumption.  


\begin{assumption}\label{assumption:completeness_z}
Let $g$ be a  square integrable function on $U$. For each $x\in\Xcal$ and for all $z\in \Zcal_p$, $
    \EE[g(U)\mid  x,z] = 0 
$ if and only if $g(U)=0$, $P(U)$ almost surely.  
\end{assumption}

Given this assumption we can prove identifiability.



\begin{proposition}\label{proposition:u}
Given that Assumptions~\ref{assumption:graph}--\ref{assumption:uz_direction}, \ref{assumption:completeness_z} 
 hold; that $m_0$ exists; that $(W,X,Y)$ are observed for the sources $z\in\Zcal_p$, and $(W,X)$ is observed from the target domain. Then $\EE_q[Y\mid x]$ is identifiable, 
and for any $x\in\Xcal$, we can write
\begin{equation}\label{eq:bridge_m0_transfer}
\EE_q[Y\mid x] = \int_{\Wcal}{m}_0(w,x)dQ(w\mid x).
\end{equation}
\end{proposition}

The proof is given in Appendix \ref{sec:multidomainContinuousProof}. Crucially, this result is valid only when Assumptions~\ref{assumption:completeness_z} 
holds, and it remains unclear when it is expected to hold. Proposition \ref{prop:numenv} suggests
that Assumptions~\ref{assumption:completeness_z} is not vacuous when $U$ is finite dimensional. We plan to investigate further this in future work. 

Now let us consider the case where $U$ is continuous. In this case, unfortunately, Assumption~\ref{assumption:completeness_z}
may not hold, preventing identification of $\EE_q[Y\mid x]$. This is illustrated in the following example. 


\begin{example}\label{example:counterexample}
Recall the decomposition of both sides of~\eqref{eq:multiDomainContinuous}.
Under Assumption \ref{assumption:CI} and given the existence of $m_0$ (Proposition \ref{prop:exsitence_m0}),
\begin{align}
\EE_{p}[{Y\mid x,z}] &= \int_{\Wcal}m_0 (w,x)\mathrm{d}P(w\mid x,z)\notag\\
&= \int_{\Ucal}\int_{\Wcal}m_0(w,x)\mathrm{d}P(w\mid u)\mathrm{d}P(u\mid x,z);\label{eq:m01}\\
\EE_{p}[{Y\mid x,z}]&= \int_{\Ucal}\EE_p[Y\mid x, u]\mathrm{d}P(u\mid x,z).\label{eq:m02}
\end{align}
For every $x$, Eqs. \eqref{eq:m01} and \eqref{eq:m02} represent projections onto $P(u\mid x,z_r),$ $r\in{1,\ldots , k_z}.$ 
Consider  $\Ucal:=[-\pi,\pi]$ with periodic boundary conditions, and for a given $x$ define $P(u\mid x,z_r) = (2\pi)^{-1} (1+\cos (r u)),\forall r\in \mathbb{N}^+$ (note that cosines form an  orthonormal basis).  
We now construct an example where (\ref{eq:m01}) holds for some $z$ but not for others. Define the difference
\begin{align}
&\EE_p[Y\mid x, u] - \int_{\Wcal}m_0(w,x)\mathrm{d}P(w\mid u) \label{eq:differenceNeeded} \\
&=
\cos ((k_z +1)u)
=:g(u). \nonumber 
\end{align}
In this case, $g(u)\neq0,$ and in particular, \eqref{eq:m01} holds for all $r\le k_z,$ but not for $P(u\mid x,z_{k_z+1}).$
\end{example}
This example illustrates a larger point: that for continuous $U$, no finite set  of projections  will suffice to completely characterize the square integrable functions on $\Ucal$. That said, as more projections are employed, and subject to appropriate assumptions on the smoothness of (\ref{eq:differenceNeeded}), the error will reduce as more domains are observed. The characterization of this convergence will be the topic of future work.  In  experiments, we show that the adaptation can still be effective even when the latent variable $U\vert z_{r}$ is continuous valued and follows different Beta distributions for each distinct $r$,  given just two training source domains. 
\vspace{-0.3cm}
\section{Kernel Bridge Function Estimation}\label{sec:kernelapproach}
\vspace{-0.3cm}

We introduce kernel methods to estimate the bridge functions and subsequently leverage the estimates to adapt to distribution shifts. Section~\ref{ssec:identification} shows that bridge functions for both settings can be adapted to the target domain, so we drop the domain specific indices and use $h_0$ and $m_0$ to denote the bridge functions. 
We begin by introducing the notation.

\textbf{Notation.} Let $\otimes$ be the tensor product, $\overline{\otimes}$ be the columnwise Khatri-Rao product and $\odot$ be the Hadamard product.
For any space $\Vcal\in\{\Xcal, \Ccal, \Wcal,\Ycal\}$, let $k:\Vcal\times\Vcal\rightarrow\RR$ be a positive semidefinite kernel function and
$\phi(v)=k(v,\cdot)$ for any $v\in\Vcal$ be the feature map. We denote $\Hcal_\Vcal$ to be the RKHS on $\Vcal$ associated with kernel function $k$. The RKHS has two properties: (i) $f\in\Hcal_{\Vcal}$, $f(v)=\dotp{f}{k(v,\cdot)}$ for all $v\in\Vcal$ and (ii) $k(v,\cdot)\in\Hcal_\Vcal$. We denote $\dotp{\cdot}{\cdot}$ as the inner product and $\opnorm{\cdot}{\Hcal_\Vcal}$ as the induced norm. 
For notation simplicity, we denote the product space $\Hcal_{\Vcal}\times\Hcal_{\Vcal'}$ associated with operation $\Hcal_{\Vcal}\otimes\Hcal_{\Vcal'}$ as $\Hcal_{\Vcal\Vcal'}$. We define the kernel mean embedding as $\mu_V=\EE[\phi(V)]=\int k(v,\cdot)p(v)dv$~\citep{smola2007hilbert} and the conditional mean embedding as $\mu_{V\mid y}=\int k(v,\cdot)p(v\mid y)dv$~\citep{song2009hilbert, singh2019kernel}. For $V\in\{W,X,C\}$, we denote the $a$-th batch of \emph{i.i.d.} samples as $V_a=\{v_{a,i}\}_{i=1}^{n_a}$. Define the Gram matrices as $\Kcal_{V_a}=\begin{bmatrix}k(v_{a,i}, v_{a,j})\end{bmatrix}_{i,j}\in\RR^{n_a\times n_a}$, $\Kcal_{V_{ab}}=\begin{bmatrix}k(v_{a,i}, v_{b,j})\end{bmatrix}_{i,j}\in \RR^{n_a\times n_b}$. Let $\Phi_{V_a}=\begin{bmatrix}\phi(v_{a,1}),\ldots,\phi(v_{a,n_a})\end{bmatrix}^\top\in\Hcal_{\Vcal}^{n_a}$ be the vectorized feature map such that
$\Phi_{V_a}(v')=\begin{bmatrix}k(v_{a,1}, v'),\ldots, k(v_{a,n_a}, v')\end{bmatrix}^\top\in\RR^{n_a}$.

\vspace{-0.1cm}
\subsection{Adaptation with Concepts}\label{ssec:estimator_h0m0} \vspace{-0.1cm}
Suppose that for the bridge function $h_0\in\Hcal_{\Wcal\Ccal}$, where  $\Hcal_{\Wcal\Ccal}$ is a RKHS. It follows from Theorem~\ref{theorem:complete_identification} that
    \begin{align}
        \EE_q[Y\mid X=x]&=\EE_q[h_0(W,C)\mid x]\notag\\
        &=\EE_q[\dotp{h_0}{\phi(W)\otimes \phi(C)}\mid x]\notag\\
        &=\dotp{h_0}{\mu_{WC\mid x}^q}.\label{eq:full_e}
    \end{align}
    
To adapt to the distribution shifts, we estimate the bridge function $h_0$ in the source domain and the conditional mean embedding $\mu_{WC\mid x}^q=\EE_q[\phi(W)\otimes\phi(C)\mid x]$ in the target domain. The empirical estimate of the conditional mean embedding along with the consistency proof have been provided in~\citep{song2009hilbert, grunewalder2012conditional} thus we focus on the estimation procedure of the bridge function $h_0$. 

To estimate the bridge function $h_0$, we employ the regression method developed in~\citet{mastouri2021proximal}. 
Recall $\EE[Y\mid c, x]=\EE[h_0(W,c)\mid c, x]$. We define the population risk function in the source domain as:
\begin{align}\label{eq:risk_h}
R(h_0)&=\EE_{p}[(Y-G_{h_0}(C, X))^2];\\
G_{h_0}(x,c)
&=\dotp{h_0}{\mu_{W\mid c, x}^p\otimes\phi(c)}.\notag
\end{align}
The procedure to optimize~\eqref{eq:risk_h} involves two stages. In the first stage, we estimate the conditional mean embedding $\mu_{W\mid c, x}^p=\EE_p[\phi(W)\mid c, x]$, which we will use as a plug-in estimator to estimate $h_0$ in the second step. Given $n_1$ \emph{i.i.d.} samples $(X_1,W_1,C_1)=\{(x_{1,i},w_{1,i},c_{1,i})\}_{i=1}^{n_1}$ from the source distribution $p$ and a regularizing parameter $\lambda_1>0$, we denote $\Kcal_{X_1}\in\RR^{n_1\times n_1}$, $\Kcal_{C_1}\in\RR^{n_1\times n_1}$ as the Gram matrices and $\Phi_{X_1}\in\Hcal_\Xcal^{n_1}$, $\Phi_{C_1}\in\Hcal_\Ccal^{n_1}$ as $n_1$-dimensional vectorized feature maps of $X_1$, $C_1$ respectively. 
Following the procedure developed in~\citet{song2009hilbert}, the estimate of $\mu_{W\mid x,c}^p$ is 
\begin{align}\label{eq:cme}
\hat{\mu}_{W\mid c,x}^p&=\sum_{i=1}^{n_1}b_i(x,c)\phi(w_{1,i});\\ b(x,c)&=(\Kcal_{X_1}\odot\Kcal_{C_1}+\lambda_1 n_1 I)^{-1}\rbr{\Phi_{X_1}(x)\odot\Phi_{C_1}(c)}.\notag
\end{align}
In the second stage, we replace $\mu_{W\mid x,c}^p$ with $\hat{\mu}_{W\mid x,c}^p$ in~\eqref{eq:risk_h} and define the empirical risk. Consider $n_2$ \emph{i.i.d.} samples $(X_2, Y_2, C_2)=\{({x}_{2,i}, {y}_{2,i},{c}_{2,i})\}_{i=1}^{n_2}$ from the source distribution and a regularization parameter $\lambda_2>0$, we want to minimize
\begin{align}\label{eq:emrisk_h}
    \argmin_{h_0\in\Hcal_{\Wcal\Ccal}} \frac{1}{2n_2}\sum_{i=1}^{n_2}\rbr{y_{2,i}-\dotp{h_0}{\phi(c_{2,i})\otimes\hat{\mu}^p_{W\mid c_{2,i}, x_{2,i}}}}^2+\lambda_2\opnorm{h_0}{\Hcal_{\Wcal\Ccal}}^2.
\end{align}
We follow the same analysis procedure derived in~\citet{mastouri2021proximal}. The solution to~\eqref{eq:emrisk_h} is shown in the following.
\begin{proposition}\label{prop:optimal_h0}
Let $\Kcal_{W_1}\in\RR^{n_1\times n_1}$, $\Kcal_{C_2}\in\RR^{n_2\times n_2}$ be the Gram matrices of $W_1$ and $C_2$, respectively. Let $\Kcal_{X_{12}}\in\RR^{n_1\times n_2}$, $\Kcal_{C_{12}}\in\RR^{n_1\times n_2}$ be the cross Gram matrices of $(X_1,X_2)$ and $(C_1,C_2)$, respectively.
For any $\lambda_2>0$, there exists a unique optimal solution to~\eqref{eq:emrisk_h} of the form 
\begin{align*}
\hat{h}_0&=\sum_{i=1}^{n_1}\sum_{j=1}^{n_2}\alpha_{ij}\phi(w_{1,i})\otimes\phi(c_{2,j});\\
\text{vec}(\alpha)&=(I\overline{\otimes} \Gamma)(\lambda_2 n_2 I +\Sigma)^{-1}y_2,
\end{align*}
where $\Sigma=(\Gamma^\top\Kcal_{W_1}\Gamma)\odot \Kcal_{C_2}$, $\Gamma=(\Kcal_{X_1}\odot\Kcal_{C_1}+\lambda_1 n_1I)^{-1}(\Kcal_{X_{12}}\odot \Kcal_{C_{12}})$, and $y_2=\begin{bmatrix} y_{2,1},\ldots, y_{2,n_2}\end{bmatrix}^\top$.
\end{proposition}
Proposition~\ref{prop:optimal_h0} is an application of the Representer theorem~\citep{scholkopf2001generalized} -- the optimal estimate of the infinite dimensional operator is a finite rank operator spanned by the feature space of $W_1$ and $C_2$. 

Finally, given estimate $\hat{\mu}_{WC\mid x}^q$ and a new sample $x_{\text{new}}$, we can construct the empirical predictor of~\eqref{eq:full_e} as
\[
\hat{y}_{\text{pred}}=\dotp{\hat{h}_0}{\hat{\mu}_{WC\mid x_{\text{new}}}^q}.
\]
This completes the full adaptation procedure. 

\textbf{On classification tasks}. For classification tasks, where the label is $Y\in\{1,\ldots, k_Y\}$, we treat the multi-task regressor as a classifier. We encode $Y$ by a one-hot encoder and then regress on the encoded $\tilde{Y}\in\{0,1\}^{k_Y}$.  Each label $\ell$ has a corresponding bridge function $h_{0,\ell}$ for $\ell\in\{1,\ldots,k_Y\}$. For $i=1,\ldots,n_2$, let the encoded $y_{2,i}$ be $\tilde{y}_{2,i}=\begin{bmatrix}\tilde{y}_{2,i,1},\ldots, \tilde{y}_{2,i,k_Y}\end{bmatrix}^\top\in\{0,1\}^{k_Y}$.  Then for each $\ell$, we can estimate $h_{0,\ell}$ by replacing  $y_{2,i}$ in~\eqref{eq:emrisk_h} with $\tilde{y}_{2,i,\ell}\in\{0,1\}$. For each new sample $x_{\text{new}}$, the predicted score  of label $\ell$ is $\hat{y}_{\text{pred},\ell}=\dotp{\hat{h}_{0,\ell}}{\hat{\mu}^q_{WC\mid x_{\text{new}}}}$, and we select the label that has the highest prediction score: $\argmax_{\ell}\hat{y}_{\text{pred},\ell}$.

\vspace{-0.2cm}
\subsection{Adaptation with Multiple Domains}
\vspace{-0.2cm}
In the multiple source domain setting, the estimation of $m_0$ follows similarly to that of $h_0$.  Assuming that $m_0\in\Hcal_{\Wcal\Xcal}$,  then~\eqref{eq:multiDomainContinuous} can be  written as
\[
\EE_r[Y\mid x]=\EE_p[\dotp{m_0}{\mu_{W\mid x,r}\otimes \phi(x)}\mid x],
\]
for $r=1,\ldots, k_Z$. 
The task is to estimate $m_0$ from the source domain and then apply it to the target domain. We can define the population risk function as
\begin{align}\label{eq:risk_m}
R(m_0)&=\sum_{r=1}^{k_Z}\EE_r[(Y-G_{m_0}(r, X))^2];\\
G_{m_0}(r,x)&=\dotp{m_0}{\mu_{W\mid r, x}\otimes\phi(x)}.\notag
\end{align}
We employ the two-stage estimation procedure as we did for estimating $h_0$: (i) we first estimate $\mu_{W\mid r, x}$ and then (ii) plug the estimate $\hat{\mu}_{W\mid r, x}$ to estimate $m_0$.



At the $r$-th domain, we observe the samples: $\{(w_{r,i}, x_{r, i}, r)\}_{i=1}^{n_{r}}$ .  As with~\eqref{eq:cme}, we learn a conditional mean embedding  $\hat{\mu}_{W\mid r,x}=\sum_{i=1}^{n_{r}}d_{r, i}(x)\phi(w_{r,i})$, where $d_r(x)=(\Kcal_{X_r}+\lambda_3 I)^{-1}\rbr{\Phi_{X_r}(x)}\in\RR^{n_{ r}}$ and $\lambda_3>0$ for $r=1,\ldots, k_Z$. 
In the second stage, given another batch of independent samples: $\{(y_{r,i}, x_{r,i}, r)\}_{i=1}^{n_{r}}$ for $r=1,\ldots, k_Z$, we minimize:
\begin{align}\label{eq:emrisk_m}
     \frac{1}{2\sum_{r=1}n_r}\sum_{r=1}^{k_Z}\sum_{i=1}^{n_r}\rbr{y_{r,i}-\dotp{m_0}{\phi(x_{r,i})\otimes\hat{\mu}_{W\mid r, x_{r,i}}}}^2+\lambda_4\opnorm{m_0}{\Hcal_{\Wcal\Xcal}}^2.
\end{align}
Then, $\hat{m}_0$ yields an analytical solution in similar form to $\hat{h}_0$ shown in Proposition~\ref{prop:optimal_h0} (see  Appendix~\ref{ssec:proof_kernelm0} for details). Finally, with the estimated conditional mean embedding $\hat{\mu}_{W\mid x}^q$ and a new sample $x_{\text{new}}$ from the target test set, we have
\[
\hat{y}_{\text{pred}}=\dotp{\hat{m}_0}{\hat{\mu}_{W\mid x_{\text{new}}}^q\otimes\phi(x_{\text{new}})}.
\]
We convert the regression task with $m_0$ to the classification task by learning $k_{Y}$ bridge functions, where each bridge function $m_{0,\ell}$ corresponds to label $\ell$. 
\vspace{-0.2cm}
\section{Experiments}
\vspace{-0.2cm}

\begin{figure*}
    \centering
    \begin{subfigure}[b]{0.48\textwidth}
    \includegraphics[width=\textwidth]{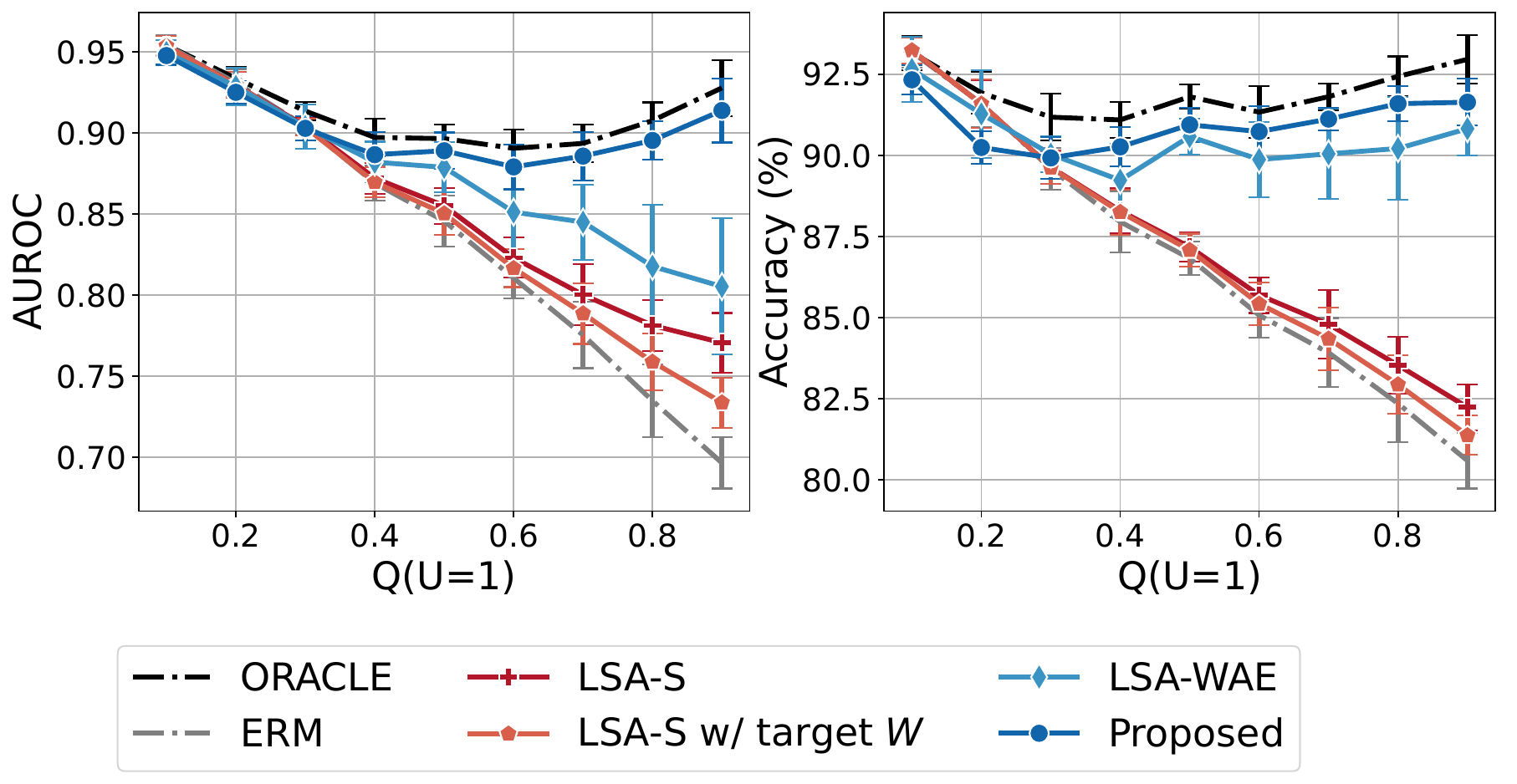}
    \vspace{-\baselineskip}
    \subcaption{Classification task on simulated data.}
    \label{fig:simu_result}
    \end{subfigure}
    \hfill
    \begin{subfigure}[b]{0.48\textwidth}
    \includegraphics[width=\textwidth]{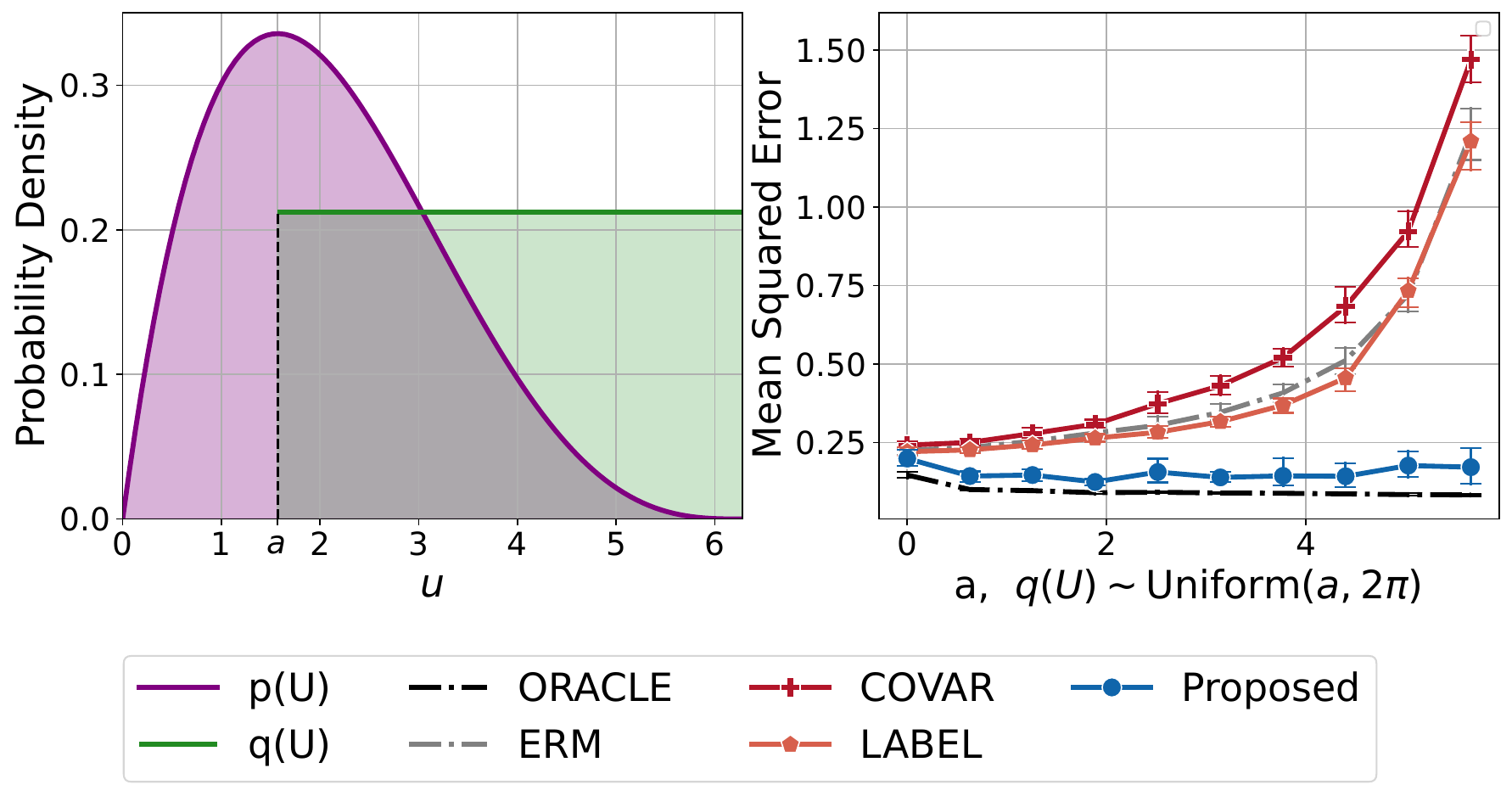}
    \vspace{-\baselineskip}
    \subcaption{Regression on the dSprites dataset.}
    \label{fig:dsprites_result}
    \end{subfigure}
    \caption{\textbf{Adaptation results with concept and proxy}. Shown is the average evaluation metric on held-out target distribution samples across 10 independent replicates of the data. The proposed method is robust to the latent shift compared to the baselines in both cases. (a) We set $P(U=1)=0.1$. Both the AUROC and accuracy remains nearly constant in various degree of shifts, while the performance of other baselines drops as $Q(U=1)$ moves to $0.9$. 
    (b) The left figure denotes the density function of $U$, the overlapping area of two distribution shrinks as $a$ moves rightward. The result on the right shows that our method is robust even when the overlapping area between two distributions is small. }
    \label{fig:main_results}
\end{figure*}

 \begin{table*}[h]
    \centering
    \caption{
    \textbf{Multi-domain adaptation result}. 
     The values are the average AUROC of $10$ independent replicates of the data. Each task has three source domains with different $P_r(U)$ and one target domain. The proposed method has outperformed other baselines and is close to the Oracle in task 2.}
    \label{tab:experiment_result}
    \resizebox{\textwidth}{!}{%

\begin{tabular}{lrrrrrrrrr}
    \toprule
    \textbf{Task} & \textbf{ORACLE} & \textbf{Cat-ERM} & \textbf{Avg-ERM} & \textbf{SA} & \textbf{MK} & \textbf{WCSC} & \textbf{DANN} & \textbf{MMD} & \textbf{Proposed} \\
    \midrule
    \multirow{2}{*}{Task 1} & $0.9425$ & $0.8030$ & $0.7916$ & $0.7918$ & $0.5848$ & $0.5221$ & $0.8039$ & $0.8055$ & $\mathbf{0.8848}$ \\
                            & $\pm0.0039$ & $\pm0.0155$ & $\pm0.0148$ & $\pm0.0148$ & $\pm0.0593$ & $\pm0.0299$ & $\pm0.0229$ & $\pm0.0248$ & $\pm0.0120$ \\
    \midrule
    \multirow{2}{*}{Task 2} & $0.9431$ & $0.8942$ & $0.8953$ & $0.8953$ & $0.8054$ & $0.8144$ & $0.9158$ & $0.9149$ & $\mathbf{0.9318}$ \\
                            & $\pm0.0061$ & $\pm0.0084$ & $\pm0.0079$ & $\pm0.0079$ & $\pm0.0204$ & $\pm0.0474$ & $\pm0.0125$ & $\pm0.0135$ & $\pm0.0063$ \\
    \midrule
    \multirow{2}{*}{Task 3} & $0.8876$ & $0.8483$ & $0.8427$ & $0.8408$ & $0.8002$ & $0.7428$ & $0.8480$ & $0.8470$ & $\mathbf{0.8569}$ \\
                            & $\pm0.0085$ & $\pm0.0134$ & $\pm0.0130$ & $\pm0.0132$ & $\pm0.0311$ & $\pm0.0311$ & $\pm0.0166$ & $\pm0.0181$ & $\pm0.0095$ \\
    \bottomrule
\end{tabular}

    }
\end{table*}

We verify our theory with both simulated and real data, demonstrating robustness to latent shifts and transferablility of the bridge functions. 

For the setting with concept variables present, we compare our method with baselines: Empricial Risk Minimization (ERM), Covariate shift weighting  (COVAR)~\citep{shimodaira2000improving}, Label shift weighting (LABEL)~\citep{buck1966comparison}, and the spectral (LSA-S) and Wasserstein Autoencoder (LSA-WAE) latent shift adaptation approaches ~\citep{alabdulmohsin2023adapting}. 
For the multi-domain setting, we compare our method with baselines: Simple Adaptation (SA)~\citep{mansour2008domain}, Weighted Combination of Source Classifiers (WCSC)~\citep{zhang2015multi}, and Marginal Kernel (MK)~\citep{blanchard2011generalizing}. We also compare with multi-domain generalization baselines~\citep{muandet2013domain}: Domain Adversarial Neural Networks (DANN) \citep{ganin2016domain}, Maximum Mean Discrepancy (MMD) \citep{GreBorRasSchetal12}. Additionally, we modify the ERM method to the multi-domain setting by concatenating the source samples to learn one ERM model (Cat-ERM) or taking the average result of each source domain ERM model (Avg-ERM).
The ORACLE model is a model that is trained on target distribution samples. and evaluated on held-out target distribution samples. The tuning parameters for all models including the proposed model are selected using five-fold cross-validation. Details regarding the setups are in Appendix~\ref{app:experiments}.


\textbf{Classification task.} The task designed in~\citet{alabdulmohsin2023adapting} is a binary classification problem with $Y\in\{0,1\}$ and the latent variable $U\in\{0,1\}$ is a Bernoulli random variable. Additionally, $X\in\RR^2, W\in\RR$ are continuous random variables and $C\in\RR^3$ is a discrete variable. We have one source domain with $P(U=1)=0.1$. We evaluate the models on the target distribution with $Q(U)$ shifting from $Q(U=1)\in\{0.1,\ldots,0.9\}$. The goal of this task is to investigate whether the adaptation method is robust to any arbitrary shift of $U$.  

The ORACLE and ERM model are implemented as MultiLayer Perceptrons (MLP). The kernel function used in the proposed method is the Gaussian kernel. 

We compare the proposed method with the LSA-S and Wasserstein Autoencoder adaptation LSA-WAE approaches developed in~\citet{alabdulmohsin2023adapting}.
While all three methods are designed to adjust shift for the same graph in Figure~\ref{fig:shiftwconcepts},
our method takes additional $W,C,X$ as training samples in the target domain while LSA-S and LSA-WAE only take $X$. 
For all three methods, only $X$ is observed in the test data. 

While the identification theory developed in~\citep{alabdulmohsin2023adapting} does not require $W,C$ in the target domain, we are aware that in practice, having more information in the target domain may improve estimation. 
To make the methods more directly comparable,  we design an additional step to incorporate $W$ from the target in the LSA-S algorithm. 
We describe this procedure in more detail in Appendix \ref{appendix:lsa_s_w}.

 
Results are shown in Figure~\ref{fig:simu_result}. The proposed method is more robust to the shift compared to baselines and is close to the ORACLE model. 
It is shown that with observed $W$ in the target domain, LSA-S does not improve the performance compared to LSA-S without $W$. We also compare results under different noise levels and observe similar trends as discussed in Appendix~\ref{app:experiments}. 

\textbf{dSprites dataset regression task.} 
We test the proposed procedure on the dSprites~\citep{dsprites17} dataset, an image dataset described by five latent parameters (shape, scale, rotation, posX, and posY). Motivated by ~\citet{dsprites17}'s experiments, we design a regression task where the dSprites images (64 $\times$ 64 = 4096-dimensional) are $X \in \RR^{64 \times 64}$ and subject to a nonlinear confounder $U \in [0, 2\pi]$ which is a rotation of the image. $W \in \RR$ and $C \in \RR$ are continuous random variables. For this experiment, we have $7000$ training samples and $3000$ test samples. 
Further details about the procedure are in Appendix~\ref{app:experiments}.

In the results in Figure~\ref{fig:dsprites_result}, we vary $a$, which controls which region of the source distribution that the target distribution concentrates. We design the experiment such that increasing $a$ shifts the target distribution to increasingly low mass regions of the source distribution. We compute the mean squared error of each method on test examples from the target distribution.

We find that, while the baseline methods degrade as the target distributions shift increases, the proposed method adapts and maintains low error, nearly matching the error achieved by the oracle, which is trained on target distribution samples.

\vspace{-0.2cm}
\subsection{Multi-Domain Adaptation}
\vspace{-0.2cm}
\begin{figure*}[ht]
    \centering
    \includegraphics[width=.9\textwidth]{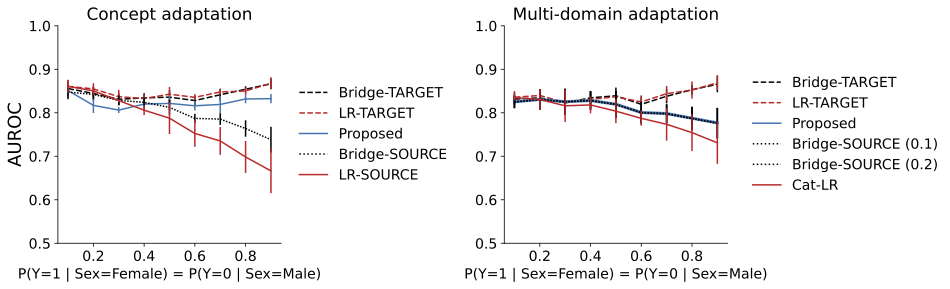}
    \caption{\textbf{Concept and multi-domain adaptation with MIMIC-CXR}. Shown are the mean $\pm$ SD AUROC of concept (left) and multi-domain adaptation (right) for classification of ``No finding'' from embeddings of chest X-rays over five replicates of a sampling procedure that introduces a shift in the prevalence of ``No finding'' with patient sex subgroups, where radiology report embeddings serve as concept variables $C$ and patient age serves as the proxy $W$. In the concept adaptation experiment, the source domain corresponds to $P(U=1) = P(Y = 1 \mid \textrm{Sex}=\textrm{Female}) = P(Y = 0 \mid \textrm{Sex}=\textrm{Male})=0.1$. In the multi-domain adaptation experiment, we consider two source domains $P(U=1)=\{0.1, 0.2\}.$ 
    }
    \label{fig:mimic_experiment}
\end{figure*}

In the multi-domain setting, we use the same classification dataset provided in~\citet{alabdulmohsin2023adapting} as Section~\ref{ssec:experiment_cp}. We assume that $C$ is not observed in any domain and generate multiple datasets drawn with different distributions on $U$. 

\textbf{Classification task.} We construct three different tasks with different settings of $P(U)$ over the source and target domains.
For each task, we construct three source domains and one target domain, drawing $3200$ random training samples for the each source domain and $9600$ random training samples for the target domain.
The set of source domains of of Task 1--3 have different combinations of distribution on $U$ documented in Appendix~\ref{ssec:classification:details}.

The backbone models for ORACLE, Cat-ERM, Avg-ERM, and SA~\citep{mansour2008domain} are simple MLPs; MK~\citep{blanchard2011generalizing} is a weighted kernel support vector machine; WCSC~\citep{zhang2015multi} is a re-weighted kernel density estimator. 
SA~\citep{mansour2008domain} assumes that $Q(X)$ is the convex combinations of $P_r(X)$ for $r=1,\ldots,k_Z$; WCSC~\citep{zhang2015multi} assumes that $Q(X\mid Y)$ is a linear mixture of $P_r(X|Y)$ for $r=1,\ldots,k_Z$ domain is an i.i.d. realization from the general distribution. 



The results are shown in Table~\ref{tab:experiment_result}.
Overall, we find our approach performs better than ERM and baseline multi-domain adaptation methods.
All methods perform better in the setting of Task 2 than for Task 1, informally demonstrating the effect of the closeness of the source domains to the target domain.
For Task 3, while our proposed approach performs best, ERM also performs well, and substantially better than the domain adaptation baselines.

\textbf{Regression task.} We consider two regression tasks, where $U$ is either a Bernoulli or a Beta random variable. We present the results in Appendix~\ref{app:experiments}. 

\subsection{Concept and multi-domain adaptation with MIMIC-CXR}

We conduct a small-scale experiment using a sample of chest X-ray data extracted from the MIMIC-CXR dataset \citep{johnson2019mimic}. We briefly describe the experimental design and results here, and include a complete description in Appendix \ref{appendix:mimic}. We consider classification of the absence of a radiological finding from low-dimensional embeddings of the X-rays \citep{sellergren2022simplified}, using the absence of a radiological finding in the radiology report as the target of prediction. This corresponds to the ``No Finding'' label defined by \citet{irvin2019chexpert}.

We consider distribution shifts similar to settings in \citet{makar2022causally}, where patient sex is considered as a possible ``shortcut" in the classification of the absence of a radiological finding. We impose distribution shift through structured resampling of the data where $P(U=1) = P(Y = 1 \mid \textrm{Sex}=\textrm{Female}) = P(Y = 0 \mid \textrm{Sex}=\textrm{Male})$ and $P(\textrm{Sex}=\textrm{Female})=P(\textrm{Sex}=\textrm{Male}) = 0.5$ is held constant. We perform both concept adaptation and multi-domain adaptation experiments with the MIMIC-CXR data. For the concept adaptation experiment, we consider the concept variable $C$ to be the embedding of a radiology report associated with the chest X-ray. We experiment with the use of patient age as a potential proxy $W$ for $U$ due to a hypothesized correlation between the presence of radiological findings and patient age. 

The results are summarized in Figure \ref{fig:mimic_experiment}. For both experiments, we find that the performance of baseline models fit using only information from the source domain(s) degrades under distribution shift. In the concept adaptation experiment, adaptation is relatively successful, as much of the performance of comparator models fit using target domain data is recovered by the adaptation procedure. 

However, we find that the multi-domain adaptation procedure is not successful. In this case, we find that while the multi-domain adaptation procedure marginally outperforms a model fit using the concatenated source domain data under distribution shift, it recovers substantially less of the performance of the target domain model than the concept adaptation procedure does. Furthermore, the adapted model does not outperform the kernel estimators that only leverage information from the source domains. The lack of success in this setting could potentially be explained by insufficient number or diversity of domains relative to the level of noise induced by sampling variability and limited sample size.

\section{Discussion}
We propose a strategy for adaptation under distribution shift in a latent variable using a bridge function approach~\citep{miao2018identifying,tchetgen2020introduction}. 
This approach allows for identification of the optimal predictor in the target domain without identifying the distribution of the latent variable and without distributional assumptions on the form of the latent.
We require that proxies of the latent variable are present and that (i) mediating concepts are available or (ii) data from multiple source domains are present.

We argue our approach is useful for two reasons.
First, the latent distribution in general is only identifiable under strict distributional assumptions~\citep{locatello2019challenging}. 
Second, recovery of the latent variable may be challenging in practice even if it is identifiable ~\citep{rissanen2021critical}. 
For example, because most latent variable estimation methods are designed to model the data generating process~\citep{kingma2013auto}, one might allocate substantial modeling capacity to variability in the data and the latent variable that are irrelevant to modeling the shift in the conditional distribution of $Y \mid X$.
By contrast, we model only the components of the observable variables relevant to the adaptation.

\FloatBarrier

{\bf Acknowledgments:} We thank Zhu Li and Dimitri Meunier for helpful discussions. AG was partly supported by the Gatsby Charitable Foundation. OS was partly supported by the UIUC Beckman Institute Graduate Research Fellowship, NSF-NRT 1735252. KT was partly supported by NSF Graduate Research Fellowship Program. SK was partly supported by the NSF III 2046795, IIS 1909577, CCF 1934986, NIH 1R01MH116226-01A, NIFA award 2020-67021-32799, the Alfred P. Sloan Foundation, and Google Inc. This study was funded by Google LLC and/or a subsidiary thereof (‘Google’).
\putbib[bu1.bbl]
\end{bibunit}

\appendix
\pagebreak
\begin{bibunit}[my-plainnat]

\section{Identification of the Distribution}

In this section, we demonstrate the existence of the bridge functions $h_0$ and $m_0$ under certain regularity conditions. We first discuss the discrete case and then generalize to the continuous case. 

\subsection{The Discrete Case of the Bridge Function $h_0$}\label{appendix:discrete_h0}

The idea of bridge function $h_0$ may seem abstract in the continuous setting. When every variable is discrete, however, the construction of the bridge function is demonstrated by solving series of matrix problems. 
 This idea originates from~\citet{miao2018identifying} and we apply the technique to show the construction of bridge function when every variable $(W,U,C,X,Y)$ is discrete. 
 
 Let 
 \begin{align*}
     \mathbf{P}(W\mid u)&=\begin{bmatrix}P(w_1\mid u)&\ldots& P(w_{k_W}\mid u)\end{bmatrix}^\top\in\RR^{k_W};\\
     \mathbf{P}(W\mid U)&=\begin{bmatrix}\mathbf{P}(W\mid u_1)&\ldots&\mathbf{P}(W\mid u_{k_U})\end{bmatrix}\in\RR^{k_W\times k_U},
 \end{align*}
  be a column vector, and a matrix, respectively. 
  We define similarly
  \begin{align*}
        \mathbf{P}(U\mid x,c)&=\begin{bmatrix}P(u_1\mid c,x)&\ldots& P(u_{k_U}\mid c, x)\end{bmatrix}^\top\in\RR^{k_U};\\
        \mathbf{P}(U\mid X, c)&=\begin{bmatrix}\mathbf{P}(U\mid x_1,c)&\ldots& \mathbf{P}(U\mid x_{k_X}, c)\end{bmatrix}\in\RR^{k_U\times k_X},
  \end{align*}
   for $c\in\Ccal$.
 We define 
 \begin{align*}
     \mathbf{P}(Y\mid X, c)&=\begin{bmatrix}\mathbf{P}(Y\mid x_1,c)&\ldots&\mathbf{P}(Y\mid x_{k_X}, c)\end{bmatrix}\in\RR^{k_Y\times k_X};\\
     \mathbf{P}(Y\mid U, c)&=\begin{bmatrix}
 \mathbf{P}(Y\mid u_1,c)&\ldots& \mathbf{P}(Y\mid u_{k_X}, c)
 \end{bmatrix}\in\RR^{k_Y\times k_X};\\
 \mathbf{P}(W\mid X,c)&=\begin{bmatrix}\mathbf{P}(W\mid x_1,c)&\ldots& \mathbf{P}(W\mid x_{k_X}, c)\end{bmatrix}\in\RR^{k_W\times k_X},
 \end{align*}
 analogously. As an alternative to finding a $h_0(w,c)$ such that
 \[
 \EE[Y\mid c,x]=\sum_{i=1}^{k_W} h_0(w_i,c)p(w_i\mid c, x), 
 \]
 the proxy problem is converted to finding a $\tilde{H}_0(Y,W,c)$ such that
 \[
 \mathbf{P}(Y\mid X, c)=\tilde{H}_0(Y, W, c)\mathbf{P}(W\mid X, c),\quad c\in\Ccal.
 \]
First, under the condition that $W\indep \{X,C\}\mid U$, we can write
 \begin{equation}\label{eq:factor_w_xc}
 \mathbf{P}(W\mid X, c) = \mathbf{P}(W\mid U)\mathbf{P}(U\mid X,c).
 \end{equation}
Similarly, under the condition that $Y\indep X\mid\{ U, C\}$, we have
\begin{equation}\label{eq:factor_y_xc}
\mathbf{P}(Y\mid X, c)=\mathbf{P}(Y\mid U, c)\mathbf{P}(U\mid X, c)
\end{equation}
We introduce the following assumption: 
\begin{assumption}\label{assumption:inverse} Columns of $\mathbf{P}(W\mid U)$ are linearly independent. 
For every $c\in\Ccal$, the columns of $\mathbf{P}(W\mid X, c)$ satisfy $\mathbf{P}(W\mid x, c)\in\Ncal(\mathbf{P}(W\mid U)^*)^\perp$ for all $x\in\Xcal$. 
\end{assumption}
Assumption~\ref{assumption:inverse} is the requirement for the least-squares problem to have an unique solution.
Hence, by Assumption~\ref{assumption:inverse}, we have
\[
\mathbf{P}(U\mid X, c) = \mathbf{P}(W\mid U)^{\dagger}\mathbf{P}(W\mid X,c),
\]
where $\mathbf{P}(W\mid U)^{\dagger}$ is the generalized inverse of $\mathbf{P}(W\mid U)$.
Plug the above equation into~\eqref{eq:factor_y_xc}, we see that
\[
\mathbf{P}(Y\mid X, c)=\underbrace{\mathbf{P}(Y\mid U, c)\mathbf{P}(W\mid U)^{\dagger}}_{\tilde{H}(Y, W, c)}\mathbf{P}(W\mid X,c).
\]


\subsection{Existence of the Bridge Function $h_0$}\label{appendix:existence:h_0}
The sufficient conditions of existence of $h_0$ are originally discussed in~\citet{miao2018identifying}, we adapt them to our setting and provide a brief review in this section. We assume the following completeness assumption and regularity conditions. This assumption is equivalent to Condition (iii) in~\citet{miao2018identifying}. 

\begin{assumption}\label{assumption:completeness_h0}
For any mean squared integrable function $g$ and for $c\in\Ccal$, $\EE[g(X)\mid W, c]=0$ almost surely if and only if $g(X)=0$ almost surely.
\end{assumption}
Let $f$ be either the distribution from $p$ or $q$, we consider $K_c:L_2(W\mid c)\rightarrow L_2(X\mid c)$ as the conditional expectation operator associated with the kernel function
\[
k(w,x,c)=\frac{f(w,x\mid c)}{f(w\mid c)f(x\mid c)}.
\]
Then it follows that $\EE[Y\mid c,x]=K_ch_0$:
\begin{align*}
    \EE[Y\mid c,x] &= \int_{\Wcal} h_0(w,c)f(w\mid x,c)\mathrm{d}w\\
    &=\int k(w,x,c)h_0(w,c)f(w\mid c)\mathrm{d}w= K_ch_0. 
\end{align*}
To find the solution $h_0$, we assume the followings.
\begin{assumption}\label{assumption:rc_h0} For any $c\in\Ccal$, 
$\int_{\Wcal}\int_{\Xcal} f(w\mid c,x)f(x\mid c, w)\mathrm{d}w\mathrm{d}x<\infty$.
\end{assumption}
This is a sufficient condition to ensure that $K_c$ is a compact operator~\citep[Example~2.3]{carrasco2007linear}. Hence, by the definition of a compact operator, there exists a singular system $\{\lambda_{c,i},\phi_{c,i},\psi_{c,i}\}_{i\in\NN}$ of $K_c$ for every $c\in\Ccal$. 
\begin{assumption}\label{assumption:rc_h0_2}
 For fixed $c\in\Ccal$:
\begin{enumerate}
    \item $\EE[Y\mid X, c]\in L_2(X\mid c);$
    \item $\sum_{i\in\NN}\lambda_{c,i}^{-2}\abr{\dotp{\EE[Y\mid X, c]}{\psi_{c,i}}}^2<\infty$.
\end{enumerate}
\end{assumption}
The above two assumptions are restatements of Conditions~(v)--(vii) in~\citet{miao2018identifying}. We adapt the results from Proposition~1 in~\citet{miao2018identifying} to the graph in Figure~\ref{fig:shiftwconcepts} which replaces the node $X$ by $C$ and node $Z$ by $X$. 
\begin{proposition}[Existence of $h_0$, adapted from Proposition~1 in~\citet{miao2018identifying}]
\label{prop:exsitence_h0} Under Assumption~\ref{assumption:CI}, \ref{assumption:completeness_h0}--\ref{assumption:rc_h0_2}, the solution to~\eqref{eq:h0} exists. 
\end{proposition}

\begin{proof}
The proof follows directly from the result of Picard's theorem. Assumption~\ref{assumption:rc_h0} implies that $K_c$ is a compact operator. Assumption~\ref{assumption:completeness_h0} implies that $\Ncal(K_c^*)^\perp=L_2(X\mid c)$. Therefore, under the first statement in Assumption~\ref{assumption:rc_h0_2}, we have $\EE[Y\mid X, c]\in\Ncal(K_c^*)^\perp$. Along with the second statement in Assumption~\ref{assumption:rc_h0_2}, we can apply Lemma~\ref{lemma:picard}.
\end{proof}

\subsection{Existence of Bridge Function $m_0$}
\label{appendix:existence:m_0}
The proof of the existence of $m_0^p$ is similar to the analysis of $h_0$. 
Let $K_x:L_2(W\mid x)\rightarrow L_2(Z\mid x)$ be  the integral operator associated with the kernel function $k(w,x,z)=p(w,z\mid x)/(p(w\mid x)p(z\mid x))$. Then, we can write 
\begin{align*}
\EE_p[Y\mid x,z]&=\int k(w,x,z)p(w\mid x)m_0(w,x)\mathrm{d}w=K_xm_0.
\end{align*}
\begin{proposition}[Existence of $m_0$, Proposition~1 in~\citet{miao2018identifying}]\label{prop:exsitence_m0} 
Assume that 
\begin{enumerate}
    \item for any mean squared integrable function $g$ and for $x\in\Xcal$, $\EE[g(Z)\mid W, x]=0$ almost surely if and only if $g(Z)=0$ almost surely;
    \item For any $x\in\Xcal$, $\int_{\Wcal}\int_{\Zcal} f(w\mid x,z)f(z\mid x, w)\mathrm{d}w\mathrm{d}z<\infty$; 
    \item For any $x\in\Xcal$, $\EE[Y\mid Z,x]\in L_2(Z\mid x)$;
    \item For any $x\in\Xcal$, $\sum_{i\in\NN}\lambda_{x,i}^{-2}\abr{\dotp{\EE[Y\mid Z, x]}{\psi_{x,i}}}^2<\infty$, where $(\lambda_{x,i},\phi_{x,i},\psi_{x,i})$ is the singular system of $K_x$. 
\end{enumerate}
Then the solution of $m_0^p$ exists.
\end{proposition}
The proof of Proposition~\ref{prop:exsitence_m0} is similar to the proof of Proposition~1 in~\citep{miao2018identifying}, where we replace $P(y | z, x)$ in Proposition~1 of~\citet{miao2018identifying} with $\EE[Y\mid Z,x]$. The proof for existence of $m_0^q$ also follows similarly as Proposition~\ref{prop:exsitence_m0}. 

\subsection{Auxiliary Lemma}
We introduce the Picard's theorem as follows.
\begin{lemma}[Picard's Theorem]\label{lemma:picard}
Let $K:\Hcal_1\rightarrow\Hcal_2$ be a compact operator with singular system $\{\lambda_j,\varphi_j,\psi_j\}_{j=1}^\infty$ and $\phi$ be a given function in $\Hcal_2$. Then the equation of first kind $Kh=\phi$ have solutions if and only if
\begin{enumerate}
    \item $\phi\in\Ncal(K^*)^\perp$, where $\Ncal(K^*)=\{h:K^*h=0\}$ is the null space of the adjoint operator $K^*$.
    \item $\sum_{j=1}^{+\infty}\lambda_j^{-2}\abr{\dotp{\phi}{\psi_j}}^2<\infty$.
\end{enumerate}
\end{lemma}

\section{Transferring Bridge Functions}

In this section, we discuss the identifiability results.

\subsection{Proof of Theorem~\ref{theorem:complete_identification}}\label{ssec:theorem:concept}
For $f\in\{p,q\}$, recall that
\begin{align*}
    \EE_f[Y\mid c,x]&=\int_{\Wcal} h_0^f(w,c)f(w\mid c,x)\mathrm{d}w\\
    &=\int_{\Wcal}\int_{\Ucal} h_0^f(w,c)f(w\mid c,u)f(u\mid c,x)\mathrm{d}u\mathrm{d}w\\
    &=\int_{\Wcal}\int_{\Ucal} h_0^f(w,c)f(w\mid u)f(u\mid c,x)\mathrm{d}u\mathrm{d}w&(W\indep C\mid U).
\end{align*}
Similarly, we can write
\begin{align*}
\EE_f[Y\mid c,x]&=\int_{\Ucal}\EE_f[Y\mid c,u]f(u\mid c,x)\mathrm{d}u&(Y\indep X\mid \{U,C\}).
\end{align*}
Under Assumption~\ref{assumption:completeness}, we have
\begin{equation}
\EE_f[Y\mid c,U]=\int_{\Wcal}h_0^f(w,c)f(w\mid U)\mathrm{d}w\quad \label{eq:tmp1}
\end{equation}
almost surely with respect to $F(U)$,  $F\in\{P,Q\}$. 

Suppose that 
$u\in\Ucal$ such that $Q(u)>0$. 
Then, by Assumption~\ref{assumption:subsetsupport} 
, we must have $P(u)>0$. Hence, conditioned on the selected $u$ and $c$ and under Assumption~\ref{assumption:graph}, we have

\begin{align*}
    \EE_p[Y\mid c,u]&=\int_{\Wcal}h_0^p(w,c)p(w\mid u)\mathrm{d}w;\\
    \EE_q[Y\mid c,u]&=\int_{\Wcal}h_0^q(w,c)p(w\mid u)\mathrm{d}w&(p(w\mid u)=q(w\mid u),\;\forall c\in\Ccal,w\in\Wcal,u\in\Ucal).
\end{align*}

We then can write
\[
\EE_p[Y\mid c,u]-\EE_q[Y\mid c,u]=\int_{\Wcal}h_0^p(w,c)p(w\mid u)\mathrm{d}w-\int_{\Wcal}h_0^q(w,c)q(w\mid u)\mathrm{d}w.
\]
Note that, by Assumption~\ref{assumption:graph}, we have $\EE_p[Y\mid c,u]=\EE_q[Y\mid c,u]$ and hence the left hand side of the above equation is $0$ and we can conclude that:
\[
\int_{\Wcal}h_0^p(w,c)p(w\mid U)\mathrm{d}w=\int_{\Wcal}h_0^q(w,c)q(w\mid U)\mathrm{d}w
\]
$Q(U)$ almost surely. 
We complete the first part of proof. 

To show the second part of the theorem, note that we can write
\begin{align}
    \EE_q[Y\mid x]&=\EE_q[\EE_q[Y\mid C, x]\mid x]\notag\\
    &=\EE_q[\EE_q[h_0^q(W,c)\mid C,x]\mid x]\notag.\\
    \intertext{Since $p(w\mid u)=q(w\mid u)$ by Assumption~\ref{assumption:graph}, we can factorize the above equation as}
    \EE_q[Y\mid x]&=\int_{\Ccal}\sbr{\int_{\Ucal}\cbr{\int_{\Wcal}h_0^q(w,c)p(w\mid u)\mathrm{d}w}q(u\mid c,x)\mathrm{d}u}q(c\mid x)\mathrm{d}c.\notag
    &\intertext{Let the support of $U$ conditioned on $c,x$ be $\Ucal_{c,x}^1=\{u:Q(u\mid c,x)>0\}$ and $\Ucal_{c,x}^0=\{u:Q(u\mid c,x)=0\}$. Hence, we have $\Ucal=\Ucal_{c,x}^0\cup\Ucal_{c,x}^1$, and $\Ucal_{c,x}^0\cap\Ucal_{c,x}^1=\emptyset$ such that $\int_{\Ucal_{c,x}^0}q(u\mid c,x)\mathrm{d}u=0$ and $\int_{\Ucal_{c,x}^1}q(u\mid c,x)\mathrm{d}u=1$. Then, we can further decompose the above as}
    \EE_q[Y\mid x]&=\int_{\Ccal}\sbr{\int_{\Ucal_{c,x}^0}\cbr{\int_{\Wcal}h_0^q(w,c)p(w\mid u)\mathrm{d}w}q(u\mid c,x)\mathrm{d}u}q(c\mid x)\mathrm{d}c\notag\\
    &\quad+\int_{\Ccal}\sbr{\int_{\Ucal_{c,x}^1}\cbr{\int_{\Wcal}h_0^q(w,c)p(w\mid u)\mathrm{d}w}q(u\mid c,x)\mathrm{d}u}q(c\mid x)\mathrm{d}c\notag\\
    &=\int_{\Ccal}\sbr{\int_{\Ucal_{c,x}^1}\cbr{\int_{\Wcal}h_0^q(w,c)p(w\mid u)\mathrm{d}w}q(u\mid c,x)\mathrm{d}u}q(c\mid x)\mathrm{d}c.\notag
    \intertext{
    Given $c,x$, since the support of $Q(U\mid c,x)$ is included in the support of $Q(U)$, so if $u\in\Ucal_{c,x}^1$, we must have $Q(u)>0$ and hence $P(u)>0$ by Assumption~\ref{assumption:subsetsupport}, and we can swap $h_0^q$ with $h_0^p$. 
    }
    &=\int_{\Ccal}\sbr{\int_{\Ucal_{c,x}^1}\cbr{\int_{\Wcal}h_0^p(w,c)p(w\mid u)\mathrm{d}w}q(u\mid c,x)\mathrm{d}u}q(c\mid x)\mathrm{d}c.\notag
    \intertext{Since $\int_{\Ucal_{c,x}^0}\cbr{\int_{\Wcal}h_0^p(w,c)p(w\mid u)\mathrm{d}w}q(u\mid c,x)\mathrm{d}u=0$, we can add it to the above term and arrive at}
    &= \int_{\Ccal}\sbr{\int_{\Ucal}\cbr{\int_{\Wcal}h_0^p(w,c)p(w\mid u)\mathrm{d}w}q(u\mid c,x)\mathrm{d}u}q(c\mid x)\mathrm{d}c\notag\\
    &=\int_{\Ccal}\int_{\Wcal}h_0^p(w,c)q(w,c\mid x)\mathrm{d}w\mathrm{d}c.\label{eq:main_identification}
\end{align}
Since we can identify $h_0^p$ from the observable $(W,X,Y,C)$ of the source domain by solving the linear system~\eqref{eq:h0}, given observable $(W,C,X)$ from the target domain, we can identify $\EE_q[Y\mid x]$.

\subsection{Proof of Proposition~\ref{prop:numenv}}\label{ssec:prop:numenv}

The following proof is a generalization of the proof of~\citet{miao2018identifying}, suited to the multidomain case.
All variables besides $Z$
are assumed to be discrete-valued and multivariate: $V$ can take $k_{v}$ values
for $V\in\{U,X,Y,W\}$.


Let $\mathbf{P}(W\mid U)=\begin{bmatrix}\mathbf{P}(W\mid u_1)&\ldots& \mathbf{P}(W\mid u_{k_U})\end{bmatrix}\in\RR^{k_W\times k_U}$. 

Similarly, define  
$$
\mathbf{P}(Y\mid U,x)=\begin{bmatrix}\mathbf{P}(Y\mid u_1,x)&\ldots&\mathbf{P}(Y\mid u_{k_U},x)\end{bmatrix}\in\RR^{k_Y\times k_U}.$$ 
This notation carries through to the remaining variables. 

 The approach we will take differs from the concept case (and standard
proxy case) in the following way: we do not observe $Z$ in the training
or test domains, nor do we know its true dimension (indeed $Z$ may
be continuous valued). Rather, we assume that we have at least $k_{Z}$
distinct draws $z_{r}$ from $Z$ in training, where $r\in\{1,\ldots,k_{Z}\}$
is the domain index, and that $k_{Z}\ge k_{U}.$ We also suppose that
in test, we observe a distinct draw $z_{k_{Z}+1}$ which was not seen
in training.

Our goal is to obtain a bridge function, which in the categorical case will be a bridge {\em matrix} of dimension $M_{w,x}\in \RR^{k_Y\times k_W}$. Define $P_{r}(V\mid x):=P(V\mid x,z_{r})$ for $V\in\{U,Y,W\}$.
We assume that for each $x$, 

\[
\mathrm{rank}\left(P_{1:k_{Z}}(U\mid x)\right)=k_{U},\qquad P_{1:k_{Z}}(U\mid x):=\left[\begin{array}{ccc}
P_{1}(U\mid x) & \hdots & P_{k_{Z}}(U\mid x)\end{array}\right]
\]
which implies that $P(U\mid x,z_{r})$ varies with $z_{r},$ and
that we see a sufficient diversity of domains to span the space of
vectors on $U$. 


The graphical model supports the conditional independence relation
\[
\{Y,X,W\}\indep Z\mid U,
\]
however we will only require the standard proxy assumptions
\begin{align*}
 & W\indep X,Z\mid U,\\
 & Y\indep Z\mid X,U.
\end{align*}


Next, as in the concept case, we require
\[
P(Y|U,x)=M_{w,x}P(W|U),
\]
where we assume $\mathrm{rank}(P(W|U))=k_{u}$ (as in the first condition of Assumption \ref{assumption:inverse}). The matrix $M_{w,x}$
is invariant to the distribution $P(U)$ by construction. If we can
solve for $M_{w,x}$, then given a novel domain corresponding to the
draw $z_{k_{z}+1}$, we have
\begin{align*}
P(Y|U,x)P_{k_{z}+1}(U\vert x) & =M_{w,x}P(W|U)P_{k_{z}+1}(U\vert x)\\
P_{k_{z}+1}(Y|x) & =M_{w,x}P_{k_{z}+1}(W|x).
\end{align*}
This allows us to compute conditional expectations under  $P(Y\mid x)$ in the novel domain,
based on observations of $(W,X)$ in this domain.

To solve for $M_{w,x}$, we project both sides on a basis over $U$
arising from the training domains,
\begin{align*}
P(Y|U,x)P_{1:k_{Z}}(U\mid x) & =M_{w,x}P(W|U)P_{1:k_{Z}}(U\mid x),
\end{align*}
where we define $P_{1:k_{Z}}(Y|x)=\left[\begin{array}{ccc}
P_{1}(Y\mid x) & \hdots & P_{k_{Z}}(Y\mid x)\end{array}\right]$, and likewise $P_{1:k_{Z}}(W\mid x).$ Then the above becomes
\begin{align}
P_{1:k_{Z}}(Y|x) & =M_{w,x}P_{1:k_{Z}}(W\mid x)\nonumber \\
M_{w,x} & =P_{1:k_{Z}}(Y|x)P_{1:k_{Z}}^{\dagger}(W\mid x).\label{eq:domainInvariantMultiDiscrete}
\end{align}
This demonstrates that we can recover the domain-invariant $M_{w,x}$
purely from observed data.

\textbf{One domain is not enough:} We illustrate with an example,
where we again consider the case where all variables are categorical: 
\begin{equation}
P(Y|x)=M_{w,x}P(W|x),\label{eq:goalEquation-1-1}
\end{equation}
where $P(Y\mid x)$ is a $k_{Y}\times1$ vector of probabilities, $P(W\mid x)$
is a $k_{W}\times1$ vector of probabilities, and $M$ is a $k_{Y}\times k_{W}$
matrix for which we wish to solve. We have too few equations for the
number of unknowns. 

One solution to \eqref{eq:goalEquation-1-1} is the matrix of conditional
probabilities $M_{w,x}=P(Y|W,x)$. This matrix is \emph{not} invariant
to changes to $P(U)$, however:
\[
p(Y|W,x)=p(Y|U,x)P(U|W,x).
\]
The posterior $P(U|W,x)$ changes when the prior $P(U)$ changes.
In contrast, the solution in \eqref{eq:domainInvariantMultiDiscrete}
is guaranteed to be domain invariant.

\subsection{Proof of Proposition~\ref{proposition:u}}\label{sec:multidomainContinuousProof}

For all $r=1,\ldots, k_Z$, we can write
\begin{align}
\EE_r[Y\mid x]=\EE[{Y\mid x,z_r}] &= \int_{\Wcal}m_0(w,x)\mathrm{d}P(w\mid x,z_r)\notag\\
&= \int_{\Ucal}\int_{\Wcal}m_0(w,x)\mathrm{d}P(w\mid u)\mathrm{d}P(u\mid x,z_r);\label{eq:m01_append}\\
\EE[{Y\mid x,z_r}]&= \int_{\Ucal}\EE[Y\mid x, u]\mathrm{d}P(u\mid x,z_r).\label{eq:m02_append}
\end{align}
By Assumption~\ref{assumption:completeness_z}, the integrands of~\eqref{eq:m01_append}--\eqref{eq:m02_append} have the following property
\begin{align}\label{eq:temp}
\EE[Y\mid x, u] = \int_{\Wcal}m_0(w,x)\mathrm{d}P(w\mid u),
\end{align}
almost surely with respect to $P(U)$. 
We will show that $m_0$ can be transferred to identify the distribution in the target domain.

We define the support set $\Scal_q(x)=\{u:Q(u\mid x)>0\}$. 
Therefore, we can write
\begin{align}
    \EE_{q}[Y\mid x] 
&=\int_{\Ucal}\EE[Y\mid u, x]\mathrm{d}Q(u\mid x)\notag\\
&=\int_{\Scal_q(x)}\EE[Y\mid u, x]\mathrm{d}Q(u\mid x)\notag.
\end{align}
 Furthermore, since we have $\Scal_q(x)\subseteq \{u:P(u)>0\}$, we can apply~\eqref{eq:temp} to obtain
\begin{align*}
\EE_{q}[Y\mid x] &={\int_{\Wcal}\int_{\Ucal}m_0(w,x)\mathrm{d}P(w\mid u)}\mathrm{d}Q(u\mid x)\\
    &= \EE_q[{m}_0(W,x)\mid x].
\end{align*}
We complete the proof.

\section{Estimation Procedure}
The estimation procedure of $\hat{h}_0$ is discussed in Section~\ref{ssec:prop:optimal:h0} and the estimation procedure of $\hat{m}_0$ is discussed in Section~\ref{ssec:proof_kernelm0}. In Section~\ref{ssec:discretezc}, we discuss the case when either $Z$ or $C$ is a discrete variable. 

\subsection{Proof of Proposition~\ref{prop:optimal_h0}}\label{ssec:prop:optimal:h0}
The proof of Proposition~\ref{prop:optimal_h0} simply follows the result in~\citep{mastouri2021proximal} which extends from the representer theorem~\citep{scholkopf2001generalized}. There exists a $\gamma\in\RR^{n_2}$ such that
\begin{align}\label{eq:representer_h0}
    \hat{h}_0 = \sum_{j=1}^{n_2}\gamma_j \hat{\mu}_{W\mid c_{2,j}, x_{2,j}}\otimes\phi(c_{2,j}).
\end{align}
From~\citet{song2009hilbert}, we have $\hat{\mu}_{W\mid c_{2,j}, x_{2,j}}=\sum_{i=1}^{n_1}b_i(c_{2,j},x_{2,j})\phi(w_{1,i})$ and $b_i$ is the $i$-th element of $b$, a function on $\Ccal\times\Xcal$:  $b(c,x)=(\Kcal_{X_1}\odot\Kcal_{C_1}+\lambda_1n_1 I)^{-1}\rbr{\Phi_{X_1}(x)\odot\Phi_{C_1}(c)}$. If we expand~\eqref{eq:representer_h0} with the previous expression, we have
\[
\hat{h}_0 = \sum_{i=1}^{n_1}\sum_{j=1}^{n_2}\alpha_{ij} \phi(w_{1,i})\otimes\phi(c_{2,j}),
\]
where $\alpha_{ij}=b_i(c_{2,j},x_{2,j})\gamma_j$. Hence, the rest of the proof will focus on finding the expression of $\alpha_{ij}$. Following the proof technique developed in~\citep{mastouri2021proximal}, we introduce two following lemmas that assist the analysis. 

\begin{lemma}\label{lemma:norm_h0}
The square of the operator norm of $\hat{h}_0$, denoted as $\opnorm{\hat{h}_0}{\Hcal_{\Wcal\Ccal}}^2$, can be represented as
\[
\opnorm{\hat{h}_0}{\Hcal_{\Wcal\Ccal}}^2=\VEC(\alpha)^\top(\Kcal_{C_2}\otimes\Kcal_{W_1})\VEC(\alpha).
\]
\end{lemma}

\begin{proof}[Proof of Lemma~\ref{lemma:norm_h0}]
Write
\begin{align*}
    \dotp{\hat{h}_0}{\hat{h}_0}&=\left\langle{\sum_{i=1}^{n_1}\sum_{j=1}^{n_2}\alpha_{ij} \phi(w_{1,i})\otimes\phi(c_{2,j})},{\sum_{m=1}^{n_1}\sum_{r=1}^{n_2}\alpha_{mr} \phi(w_{1,m})\otimes\phi(c_{2,r})}\right\rangle\\
    &=\sum_{i,m=1}^{n_1}\sum_{j,r=1}^{n_2}\alpha_{ij}\alpha_{mr}k(w_{1,i},w_{1,m})k(c_{2,j},c_{2,r})\\
    &=\tr\rbr{\alpha^\top \Kcal_{W_1}\alpha \Kcal_{C_2}}\\
    &=\VEC(\alpha)^\top\VEC(\Kcal_{W_1}\alpha \Kcal_{C_2}).\\
    \intertext{Using the fact that $\VEC(ABC)=(C^\top\otimes A)\VEC(B)$, the above display can be written as}
    &=\VEC(\alpha)^\top(\Kcal_{C_2}\otimes\Kcal_{W_1})\VEC(\alpha).
\end{align*}
\end{proof}

\begin{lemma}\label{lemma:inner_prod_h0}
For any $c\in\Ccal$, $x\in\Xcal$,
\[
\dotp{\hat{h}_0}{\phi(c)\otimes\hat{\mu}_{W\mid c,x}}=\Phi_{C_2}(c)^\top \alpha^\top \Kcal_{W_1}(\Kcal_{X_1}\odot\Kcal_{C_1}+\lambda_1 n_1 I)^{-1}(\Phi_{C_1}(c)\odot\Phi_{X_1}(x)).
\]
\end{lemma}

\begin{proof}[Proof of Lemma~\ref{lemma:inner_prod_h0}]
Write
\begin{align*}
   \dotp{\hat{h}_0}{\phi(c)\otimes\hat{\mu}_{W\mid c,x}}&=
   \left\langle
   \sum_{i=1}^{n_1}\sum_{j=1}^{n_2}\alpha_{ij} \phi(w_{1,i})\otimes\phi(c_{2,j}),
   \phi(c)\otimes \sum_{r=1}^{n_1}b_r(c,x)\phi(w_{1,r})\right\rangle\\
   &=\sum_{i=1}^{n_1}\sum_{j=1}^{n_2}\sum_{r=1}^{n_1}\alpha_{ij}k(w_{1,i}, w_{1,r})k(c_{2,j},c)b_r(c,x).
   \intertext{Summing over $i,j$, the above equation is equivalent as}
   &=\sum_{r=1}^{n_1}\Phi_{C_2}(c)^\top\alpha^\top\Phi_{W_1}(w_{1,r})b_r(c,x)\\
   &=\Phi_{C_2}(c)^\top\alpha^\top\Kcal_{W_1}b(c,x)\\
   &=\Phi_{C_2}(c)^\top\alpha^\top\Kcal_{W_1}(\Kcal_{X_1}\odot\Kcal_{C_1}+\lambda_1n_1  I)^{-1}\rbr{\Phi_{X_1}(x)\odot\Phi_{C_1}(c)}\\
   &={\rbr{\Phi_{X_1}(x)\odot\Phi_{C_1}(c)}^\top(\Kcal_{X_1}\odot\Kcal_{C_1}+\lambda_1n_1  I)^{-1}\Kcal_{W_1}}\alpha\Phi_{C_2}(c).
\end{align*}

\end{proof}

With Lemma~\ref{lemma:norm_h0}--\ref{lemma:inner_prod_h0}, we can write~\eqref{eq:emrisk_h} as
\begin{align}\label{eq:vector_alpha}
    \frac{1}{2n_2}\norm{y_2-D^\top\VEC(\alpha)}_2^2+\lambda_2\VEC(\alpha)^\top E\VEC(\alpha),
\end{align}
where
\begin{align*}
    D=\Kcal_{C_2}\overline{\otimes}\cbr{\Kcal_{W_1}(\Kcal_{X_1}\odot\Kcal_{C_1}+\lambda_1n_1 I)^{-1}(\Kcal_{X_{12}}\odot\Kcal_{C_{12}})}
    ,\quad E=\Kcal_{C_2}\otimes\Kcal_{W_1}.
\end{align*}
Then by setting the gradient of~\eqref{eq:vector_alpha} with respect to $\VEC(\alpha)$ to zero, we will obtain
\begin{align}
    \VEC(\alpha)&=\rbr{DD^\top+\lambda_2n_2 E}^{-1}Dy_2.\notag
    \intertext{Apply Woodbury matrix identity, the above display is equivalent as}
    &=E^{-1}D(\lambda_2n_2I+D^\top E^{-1}D)^{-1}y_2.\label{eq:alpha2}
\end{align}
Using the fact that for matrices $A,B,C,F$,  $(A\otimes B)(C\overline{\otimes} F)=AC\overline{\otimes} BF$, we can simplify $E^{-1}D$ as
\begin{align*}
    E^{-1}D&=\rbr{\Kcal_{C_2}^{-1}\otimes{\Kcal_{W_1}^{-1}}}\sbr{\Kcal_{C_2}\overline{\otimes}\cbr{\Kcal_{W_1}(\Kcal_{X_1}\odot\Kcal_{C_1}+\lambda_1n_1 I)^{-1}(\Kcal_{X_{12}}\odot\Kcal_{C_{12}})}}\\
    &=I\overline{\otimes}(\Kcal_{X_1}\odot\Kcal_{C_1}+\lambda_1n_1 I)^{-1}(\Kcal_{X_{12}}\odot\Kcal_{C_{12}})\\
    &=I\overline{\otimes}\Gamma.
\end{align*}
Hence, using the fact that $(A\overline{\otimes} B)^\top(C\overline{\otimes} F)=(A^\top C)\odot B^\top F$, we have
\begin{align*}
    D^\top E^{-1}D= (\Kcal_{C_2}\overline{\otimes} \Kcal_{W_1}\Gamma)^\top(I\overline{\otimes}\Gamma)=\Kcal_{C_2}\odot (\Gamma^\top\Kcal_{W_1}\Gamma)
\end{align*}
Hence, we can write~\eqref{eq:alpha2} as
\[
\VEC(\alpha)=(I\overline{\otimes}\Gamma)\cbr{\lambda_2n_2I+\Kcal_{C_2}\odot (\Gamma^\top\Kcal_{W_1}\Gamma)}^{-1}y_2. 
\]
\subsection{Proof of Kernel Bridge Function $m_0$}\label{ssec:proof_kernelm0}
We begin with the results.
\begin{proposition}\label{prop:optimal_m0}
Let $\Kcal_{W_3}\in\RR^{n_3\times n_3}$, $\Kcal_{X_4}\in\RR^{n_4\times n_4}$ be the Gram matrices of $W_3$ and $X_4$, respectively. Let $\Kcal_{X_{34}}\in\RR^{n_3\times n_4}$, $\Kcal_{Z_{34}}\in\RR^{n_3\times n_4}$ be the cross Gram matrices of $(X_3,X_4)$ and $(Z_3,Z_4)$, respectively.
For any $\lambda_4>0$, there exists a unique optimal solution to~\eqref{eq:emrisk_m} of the form 
\begin{align*}
\hat{m}_0&=\sum_{i=1}^{n_3}\sum_{j=1}^{n_4}\alpha_{ij}\phi(w_{3,i})\otimes\phi(x_{4,j});\\
\text{vec}(\alpha)&=(I\overline{\otimes} \Gamma)(\lambda_4 n_4 I +\Sigma)^{-1}y_4,
\end{align*}
where $\Sigma=(\Gamma^\top\Kcal_{W_3}\Gamma)\odot \Kcal_{X_4}$, $\Gamma=(\Kcal_{X_3}\odot\Kcal_{Z_3}+\lambda_3 n_3I)^{-1}(\Kcal_{X_{34}}\odot \Kcal_{Z_{34}})$, and $y_4=\begin{bmatrix} y_{4,1},\ldots, y_{4,n_4}\end{bmatrix}^\top$.
\end{proposition}
The proof of Proposition~\ref{prop:optimal_m0} follows exactly as the proof of Proposition~\ref{prop:optimal_h0}, with $X$ replaced by $Z$ and $C$ replaced by $X$. 


\subsection{Estimation with discrete $Z$ or $C$}\label{ssec:discretezc}

In the case when $C$ or $Z$ happen to be discrete variables, a more efficient alternative to the estimator introduced in Section~\ref{ssec:estimator_h0m0} which requires kernelized features of $C$ (or $Z$), is to solve a separate regression of $W$ on $X$ for each $c\in\Ccal$ (or $z\in\Zcal$). 
Define the index set $\Xi_1(c)=\{i:c_{1,i}=c,i=1,\ldots,n_1\}$, we modify~\eqref{eq:cme} as
\begin{align*}
\hat{\mu}_{W\mid c,x}^p&=\sum_{i=1}^{n_1}b_i(x)\phi(w_{1,i})\mathds{1}(c_{1,i}=c);\\ b(x)&=(\Kcal_{X_{1,c}}+\lambda_1 I)^{-1}{\Phi_{X_{1,c}}(x)},\notag
\end{align*}
where $\Kcal_{X_{1,c}}=[k(x_{1,i},x_{1,j})]_{i,j}$ and $\Phi_{X_{1,c}}=\begin{bmatrix}\phi(x_{1,i})\end{bmatrix}_i^\top$ with $i,j\in\Xi_1(c)$. Alternatively, one can apply the form in~\eqref{eq:cme} but use binary kernel on $C$ (or $Z$). 


\section{Experiments} \label{app:experiments}
In this section we discuss the experimental settings and implementation details. We start with introducing the implementation details of all the baselines and proposed method. Then, we discuss the experimental settings. 

\subsection{Baselines of Adaptation with Concepts and Proxies}\label{appendix:lsa_s_w}
We introduce the baseline methods for the adaptation task with $C$ and $W$. This includes the baselines methods COVARS, LABELS, ORACLE, LSA-W, LSA-S, LSA-S w/ target $W$ and the proposed method. 
To select the parameters for the regression task on dSprite, we apply five-fold cross-validation with mean squared error as the metric to select the kernel length scale and the ridge regularization penalty.

\textbf{COVARS.} We fit a domain classifier using logistic regression, compute instance weights following~\citet{shimodaira2000improving}, and learn a weighted kernel ridge regressor with a Gaussian kernel function on the source training samples. 

\textbf{LABELS.} The label shift baseline assumes oracle access to labels in the target domain. For the classification task, we compute instance weights $q(Y)/p(Y)$ using the observed frequencies in the validation set for the source domain and the training set for the target domain. For the regression task, we compute the weights by fitting a Gaussian kernel density estimator using the source validation set and the target training set separately. We then use the fitted densities to estimate $q(Y)/p(Y)$ for each sample in the source training set. Finally, we learn a sample-weighted kernel ridge regressor with a Gaussian kernel on the source training samples.  

\textbf{ORACLE.} For regression tasks, we learn a kernel ridge regressor with a Gaussian kernel on target training samples. 
For the classification task, we use a standard MLP trained with sample in the target domain. Details of the model structure are documented in Section~\ref{ssec:baseline:multisource}.


\textbf{LSA-W.} The estimation procedure follows Section~6 in~\citet{alabdulmohsin2023adapting}. In this case, we discretize the values of $W$ by applying additional transform $\mathrm{sign}(w)$ for each sample $w$.

\textbf{LSA-S.} The estimation procedure follows Algorithm~2--5 in~\citet{alabdulmohsin2023adapting}.

\textbf{LSA-S w/ target $W$.} We briefly describe the procedure to incorporate target $W$ to LSA-S.  \citet{alabdulmohsin2023adapting} showed that $Q(Y|x)$ can be decomposed as
\begin{align}
Q(Y\mid x)
&=\sum_{\tilde{u}}\underbrace{P(Y\mid \tilde{u}, x)}_{(a)}\underbrace{Q(\tilde{u}\mid x)}_{(b)}\label{eq:new_formula}\\
&\propto\sum_{\tilde{u}}\underbrace{P(Y\mid \tilde{u}, x)}_{(a)}\underbrace{P(\tilde{u}\mid x)}_{(c)}\underbrace{\frac{Q(\tilde{u})}{P(\tilde{u})}}_{(d)}\frac{P(x)}{Q(x)}\label{eq:decompose_formula},
\end{align}
where $\tilde{u}$ is a permutation of original $u$. 
Both LSA-WAE and LSA-S are multi-stage procedures to compute (a), (c), (d) individually and combine the results using formula~\eqref{eq:decompose_formula} to obtain the predicted target distribution. Step (a) corresponds to Algorithm~5, (c) corresponds to Equation~(17), and (d) corresponds to Algorithm~4 in~\citep{alabdulmohsin2023adapting}.

With the additional $W$ from target, we can obtain (b) by slightly modifying the one estimation step in LSA-S. We test on this procedure, namely LSA-S w/ target W, with (c), (d) replaced by (b). Suppose that $U$ takes values in $1,\ldots,k_U$ and $\tilde{U}$ be a permutation of $U$.  Define the matrix ${\bf G}$ as:
\[
{\bf G} = \begin{bmatrix}
    \dotp{\hat{P}(W\mid\tilde{U}=1)}{\hat{P}(W\mid\tilde{U}=1)}&\cdots&
    \dotp{\hat{P}(W\mid\tilde{U}=1)}{\hat{P}(W\mid\tilde{U}=k_U)}\\
    \vdots&\ddots&\vdots\\
    \dotp{\hat{P}(W\mid\tilde{U}=k_U)}{\hat{P}(W\mid\tilde{U}=1)}&\cdots&
    \dotp{\hat{P}(W\mid\tilde{U}=k_U)}{\hat{P}(W\mid\tilde{U}=k_U)}    
    \end{bmatrix},
\]
where $\hat{P}(W\mid\tilde{U}=i)$ is the estimated conditional kernel density function obtained by Algorithm~3 in~\citet{alabdulmohsin2023adapting}. 
The step (b) is computed by solving the following least-squares:
\begin{align*}
    \hat{Q}(\tilde{\Ub}\mid x) =& \arg\min \left\|
    \begin{bmatrix}
    \dotp{\hat{Q}(W\mid x)}{\hat{P}(W\mid\tilde{U}=1)}\\
    \vdots\\
    \dotp{\hat{Q}(W\mid x)}{\hat{P}(W\mid\tilde{U}=k_U)}
    \end{bmatrix}
    -
    {\bf G}\begin{bmatrix}
    Q(\tilde{U}=1\mid x)\\
    \vdots\\
    Q(\tilde{U}=k_U\mid x)
    \end{bmatrix}
    \right\|_F^2,\\
    &\text{subject to }\quad 0\leq Q(\tilde{U}=i\mid x)\leq 1,\quad i=1,\ldots,k_U;\notag\\
    &\quad\quad\quad\quad\quad \sum_{i=1}^{k_U}Q(\tilde{U}=i\mid x)=1.
\end{align*}
 Then, we compute the predicted conditional probability based on~\eqref{eq:new_formula}.

\textbf{Proposed Method.} For the regression task using the dSprite dataset, we employ the Gaussian kernel function as the feature map for both $X$ and $W$. In the classification task, we also utilize the Gaussian kernel function for $X$ and $W$. Additionally, we make use of a columnwise binary kernel for $C$, which performs a binary kernel operation on each entry and computes the product of all function outputs. To compute $\hat{h}_0$, we apply one-hot encoder on $Y$ and apply the results in Proposition~\ref{prop:optimal_h0}  For choosing the kernel length scale for the classification task, we use the validation set with AUROC metric.

\subsection{Baselines of Multi-Source Adaptation}\label{ssec:baseline:multisource}

For the first three baselines: Cat-ERM, Avg-ERM, and SA, we use a \emph{standard MLP} model as the backbone structure. It is a single hidden layer MLP with size $100$ and ReLU activation functions. The network is trained using Adam optimizer~\citep{kingma2014adam} with learning rate $10^{-3}$. The batch size is set to be $200$ and the maximum number of iteration is set to be $300$.

\textbf{Cat-ERM.} We concatenate all the samples across environments into one dataset. Then, we train the model with a standard MLP model as specified above.

\textbf{Avg-ERM.} For each environment, we train a standard MLP model. During testing, we take the average of predictions from all models.

\textbf{Simple Adaptation (SA)~\citep{mansour2008domain}.}  To implement the method, we build kernel density estimators with Gaussian kernel function to estimate the density $p_r(x)$ for $r=1,\ldots, k_Z$. We then reweigh the  output of the classifier, a standard MLP, of each domain with the normalized weight $P_r(x_{\text{new}})/\cbr{\sum_{r}P_{r'}(x_{\text{new}})}$. The kernel length scale is chosen using five-fold cross-validation with AUROC metric.

\textbf{Marginal Kernel (MK)~\citep{blanchard2011generalizing}.} This method involves a kernel SVM with a product kernel on $(\Xcal,P(X))$. For any $x,x'\in\Xcal$ and a distribution on $X$, $P, P'$, the kernel function is defined as $k((x, P),(x', P'))=k_1(x,x')k_2(P,P')$. Let $n$ be the number of samples. Here $k_1$ is a Gaussian kernel function, and $k_2$ is the mean of the Gram matrix $[k(x_i,x_j')]_{ij}\in\RR^{n\times n}$, where $x_i$ for $i=1,\ldots,n$ is a \emph{i.i.d.} sample from $P$ and $x_j'$ for $j=1,\ldots,n$ is a \emph{i.i.d.} sample from $P'$.  
To accommodate the large dataset,  we precompute the Gram matrix and apply it to a linear classifier trained using Stochastic Gradient Descent (SGD) implemented in the package \emph{scikit-learn}~\citep{scikit-learn}. The kernel length scale is chosen using five-fold cross-validation with AUROC metric. 

\textbf{Weighted Combination of Source Classifiers (WCSC)~\citep{zhang2015multi}.} For each source environment, we estimate the conditional probability $X\mid y$ using kernel density estimator with the Gaussian kernel function. The rest of the estimation procedure follows Section~2 in~\citet{zhang2015multi}.  The kernel length scale is chosen using five-fold cross-validation with AUROC metric. 

\textbf{Proposed Method.} We use columnwise Gaussian kernel function as the feature map of $X$, a Gaussian kernel function as the feature map of $W$. The conditional mean embedding $\hat{\mu}_{W\mid x,z}^p$ is estimated using the approach introduced in Section~\ref{ssec:discretezc}. The analytical solution of $\hat{m}_0$ is discussed in Proposition~\ref{prop:optimal_m0}.  All the kernel length scale and the regularization parameters $\lambda_3$, $\lambda_4$ are selected using five-fold cross-validation  with AUROC metric.

\textbf{ORACLE.} The model is $\dotp{\hat{m}_0}{\hat{\mu}_{W\mid x}^q}$, where both the bridge function $\hat{m}_0$ and $\hat{\mu}_{W\mid x}^q$ are estimated using the target dataset, with the number of training samples equal to the training samples of the source domain.  All the kernel length scale and the regularization parameters $\lambda_3$, $\lambda_4$ are selected using five-fold cross-validation  with AUROC metric. 

\subsection{Classification Task} \label{ssec:classification:details}
\begin{figure}
    \centering
    \includegraphics[width=.98\textwidth]{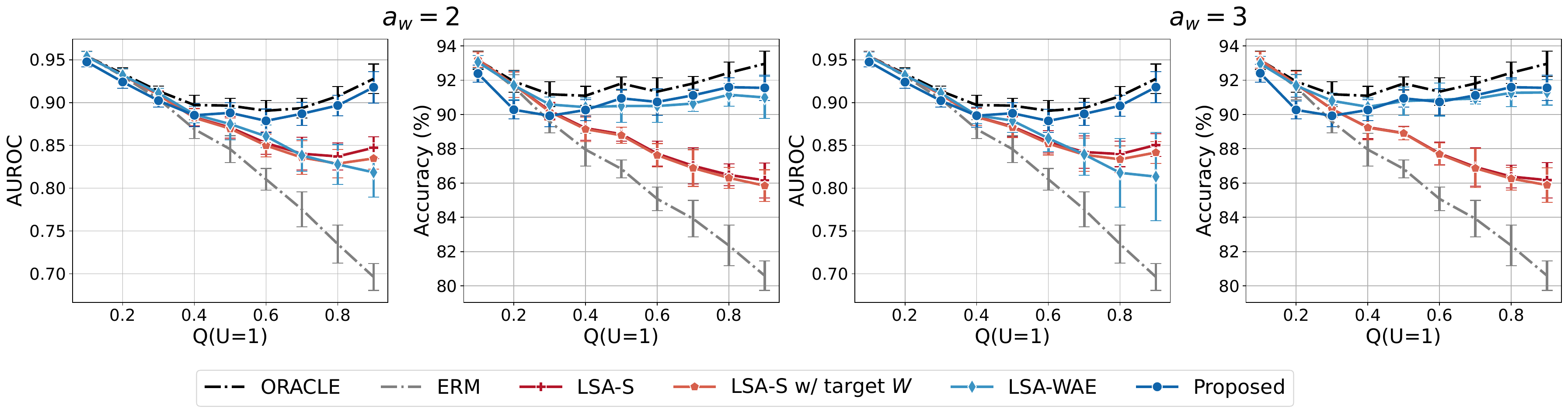}
    \caption{Classification results with $a_w=2,3$. The figures indicate that LSA-S and LSA-S w/ target $W$ have comparable performance, aggregating the target $W$ does not seem to improve the performance. }
    \label{fig:additional_results}
\end{figure}

The classification task discussed in Section~\ref{ssec:experiment_cp} is first introduced~\citet{alabdulmohsin2023adapting}. Let $\bo(\cdot)$ be the one-hot encoder, we follow their data generation procedure and generate samples using the following data generation process:
\begin{align*}
    U  \sim & \; \textrm{Categorical}(\boldsymbol{\pi}); \\
    W \mid U=u \sim & \;  \mathcal{N}(\bo(u) \mathbf{M}_{W|U}, 1 \big); \\
    X \mid U=u \sim &\; \mathcal{N}(\bo(u) \mathbf{M}_{X|U}, \mathbf{ I}_{k_X}); \\
    C_i \mid X=x,U=u \sim &\; \textrm{Bernoulli} \Big(\mathrm{logit}^{-1} \big( [x \mathbf{M}_{C|X,U=u} + \bo(u) \mathbf{M}_{C|U}]_i \big) \Big); \\
    Y \mid C=c, U=u \sim &\; \textrm{Bernoulli} \Big ( \mathrm{logit}^{-1} \big(c \mathbf{M}_{Y|C, U=u} + \bo(u) \mathbf{M}_{Y|U} \big) \Big),
\end{align*}

where the matrices are defined as
\begin{align*}
    \;& \mathbf{M}_{W|U} :=
     \begin{bmatrix}
        -1 & 1
    \end{bmatrix}^\top
    ,\quad
\mathbf{M}_{X|U} := a_w
     \begin{bmatrix}
        -1 & 1 \\
        1 & -1
    \end{bmatrix}
    ,\quad
    \mathbf{M}_{C|U} :=
    \begin{bmatrix}
        -2 & 2 & 2 \\
        -1 & 1 & 2
    \end{bmatrix}; \\
    \;& \mathbf{M}_{C|X,U=u_0} :=
    3\begin{bmatrix}
        -2 & 2 & -1 \\
        1 & -2 & -3
    \end{bmatrix} 
    ,\quad
    \mathbf{M}_{C|X,U=u_1} :=
    3\begin{bmatrix}
        2 & -2 & 1 \\
        -1 & 2 & 3
    \end{bmatrix}; \\
    \;& \mathbf{M}_{Y|U} :=
    \begin{bmatrix}
        2 & 2
    \end{bmatrix}^\top 
    ,\quad
    \mathbf{M}_{Y|C, U=u_0}  := 
    \begin{bmatrix}
        3 & -2 & -1
    \end{bmatrix} ^\top
    ,\quad
    \mathbf{M}_{Y|C, U=u_1} := 
    \begin{bmatrix}
        3 & -1 & -2
    \end{bmatrix} ^\top.
\end{align*}
The coefficient $a_w=1$ in Figure~\ref{fig:simu_result}. Figure~\ref{fig:additional_results} displays additional results where $a_w=2,3$. We generate $7000$ training samples, $1000$ validation samples, and $2000$ testing samples for the classification task with concepts and proxies.

In the multi-domain case, we construct $3$ different tasks:
Task $1$ is composed of $z_1, z_2,z_3$ such that $P(U=0\mid {z_1})=0.1$, $P(U=0\mid {z_2})=0.2$, $P(U=0\mid {z_3})=0.3$ and a target domain with $Q(U=0)=0.9$. For task $2$, we select $z_4, z_5, z_6$ such that $P(U=0\mid z_4)=0.4$, $P(U=0\mid {z_5})=0.5$, $P(U=0\mid {z_6})=0.6$ and $Q(U=0)=0.9$. For task $3$, we select $z_7, z_8, z_9$ such that $P(U=0\mid z_7)=0.7$, $P(U=0\mid {z_8})=0.8$, $P(U=0\mid {z_9})=0.9$ and $Q(U=0)=0.4$. The results are shown in Table~\ref{tab:experiment_result}--~\ref{tab:domain_generization_baselines}. 

\subsection{Comparison to Domain Generalization Baselines}
\begin{table*}[h]
    \centering
    \caption{
    \textbf{Multi-domain generalization vs. (proposed) adaptation result}. 
     The values are the average AUROC of $10$ independent runs drawn from the data generating process. Each task has three source domains with different $P_r(U)$ and one target domain. The proposed method has outperformed all domain generalization benchmarks across all tasks.}
    \resizebox{\linewidth}{!}{%

\begin{tabular}{lccccccccccc}
    \toprule
    & \textbf{ORACLE} & \textbf{ARM} & \textbf{CDANN} & \textbf{CORAL} & \textbf{DANN} & \textbf{GroupDRO} & \textbf{IRM} & \textbf{MMD} & \textbf{VREx} & \textbf{Proposed} \\
    \midrule
    \textbf{Task 1} & $0.9425$ & $0.8065$ & $0.8061$ & $0.8030$ & $0.8039$ & $0.7954$ & $0.7989$ & $0.8055$ & $0.8010$ & $\mathbf{0.8848}$ \\
    & $\pm 0.0039$ & $\pm 0.0247$ & $\pm 0.0252$ & $\pm 0.0236$ & $\pm 0.0229$ & $\pm 0.0323$ & $\pm 0.0283$ & $\pm 0.0248$ & $\pm 0.0279$ & $\pm 0.0120$ \\
    \textbf{Task 2} & $0.9431$ & $0.9143$ & $0.9159$ & $0.9158$ & $0.9158$ & $0.9160$ & $0.9131$ & $0.9149$ & $0.9136$ & $\mathbf{0.9318}$ \\
    & $\pm 0.0061$ & $\pm 0.0150$ & $\pm 0.0125$ & $\pm 0.0132$ & $\pm 0.0125$ & $\pm 0.0125$ & $\pm 0.0135$ & $\pm 0.0135$ & $\pm 0.0124$ & $\pm 0.0063$ \\
    \textbf{Task 3} & $0.8876$ & $0.8470$ & $0.8456$ & $0.8473$ & $0.8480$ & $0.8487$ & $0.8469$ & $0.8470$ & $0.8470$ & $\mathbf{0.8569}$ \\
    & $\pm 0.0085$ & $\pm 0.0171$ & $\pm 0.0164$ & $\pm 0.0163$ & $\pm 0.0166$ & $\pm 0.0185$ & $\pm 0.0186$ & $\pm 0.0181$ & $\pm 0.0132$ & $\pm 0.0095$ \\
    \bottomrule
\end{tabular}

    }
    \label{tab:domain_generization_baselines}
\end{table*}

Given that we observe multiple domains at test time, a natural question is: Does adaptation give us an advantage over generalization? In generalization, we cannot assume to have any observations in the target domain. We compare our adaptation method with multi-domain generalization baselines~\citep{muandet2013domain}: Adaptive Risk Minimization (ARM) ~\citep{zhang2021adaptive}, Conditional Domain Adversarial Neural Networks (CDANN)~\citep{long2018conditional}, Correlation Alignment (CORAL)~\citep{sun2016deep}, Domain Adversarial Neural Networks (DANN)~\citep{ganin2016domain}, Distributionally Robust Optimization for Group Shifts (GroupDRO)~\citep{sagawa2019distributionally}, Invariant Risk Minimization (IRM)~\citep{arjovsky2019invariant}, Maximum Mean Discrepancy (MMD) \citep{Borgwardt2006IntegratingSB}, and Risk Extrapolation (REx)~\citep{krueger2021out}.

In Table \ref{tab:domain_generization_baselines}, we show that our proposed method for domain adaptation in the multi-domain setting outperforms the state-of-the-art multi-domain generalization methods.

\subsection{Regression Tasks}\label{ssec:regressiontask}
We consider three tasks. We will first introduce the simulated task and then discuss about the task on dSprite data~\citep{dsprites17}.
\subsubsection{Simulated Dataset}
We consider the following data generation process. 

\begin{figure}[ht]
    \centering
    \includegraphics[width=\textwidth]{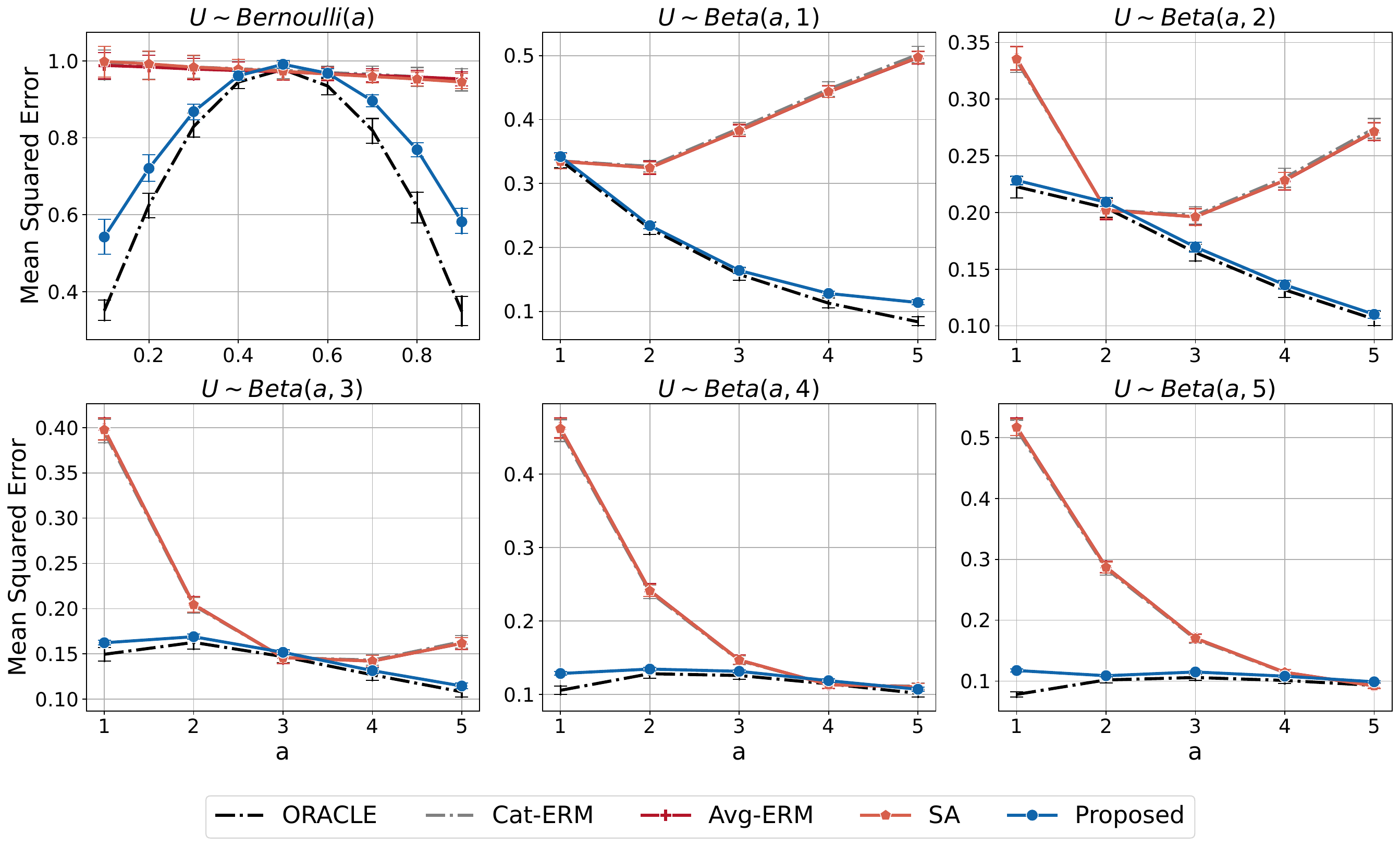}
    \caption{
    \textbf{ Top left:  results of regression task 1.}  The proposed method is close to the ORACLE method as compared all other competing methods that is vulnerable to the distribution shifts. 
    \textbf{Other figures: results of regression task 2.} In each plot, we fix $b$ and vary $a$. For all plots, it appears that when $a=b$, the mean squared error of all methods converge to a point. This is the case when the target density function of $U$ has a peak centered around $0.5$, as shown in Figure~\ref{fig:beta_distribution}, and hence $Y=(2U-1)X$ is close to zero for most samples. }
    \label{fig:beta_multisource}
\end{figure}

\textbf{Simulated regression task 1.}
\begin{align}
    U &= Ber(a);\notag\\
    X &= \Ncal(0, 1)\notag;\\
    Y &= -X\one_{\rbr{U=0}} + X\one_{\rbr{U=1}};\label{eq:gen_y}\\
    W &= \Ncal(-1, 0.01)\one_{\rbr{U=0}} + \Ncal(1, 0.01)\one_{\rbr{U=1}}\notag.
\end{align}
There are two source domains. We set $a=0.1$ for source domain $z_1$ and $a=0.9$ for source domain $z_2$. According to the data generation process~\eqref{eq:gen_y}, $Y$ is mostly positively correlated with $X$ in domain $z_1$ and negatively correlated with $X$ in domain $z_2$. For each domain, we synthesized $2000$ training samples and $1000$ testing samples. We sweep across $a=\{0.1, 0.2, 0.3, 0.4, 0.5, 0.6,0.7,0.8,0.9\}$ in the target domain. We run $10$ replications and the results shown in Figure~\ref{fig:beta_multisource}. In the next task, we set $U$ to be a continuous random variable following a Beta distribution.

In this task, we expect the Cat-ERM method to fail drastically as we anticipate that the predicted $Y$ versus $X$ is a flat line -- the predicted result would be an average of the downward sloping line $(U=0)$ and upward sloping line $(U=1)$. The result in Figure~\ref{fig:beta_multisource} supports our hypothesis, as the mean squared error remains nearly  flat  as we vary the target distribution $Q(U)$.

\begin{figure}[ht]
    \centering
    \includegraphics[width=.8\textwidth]{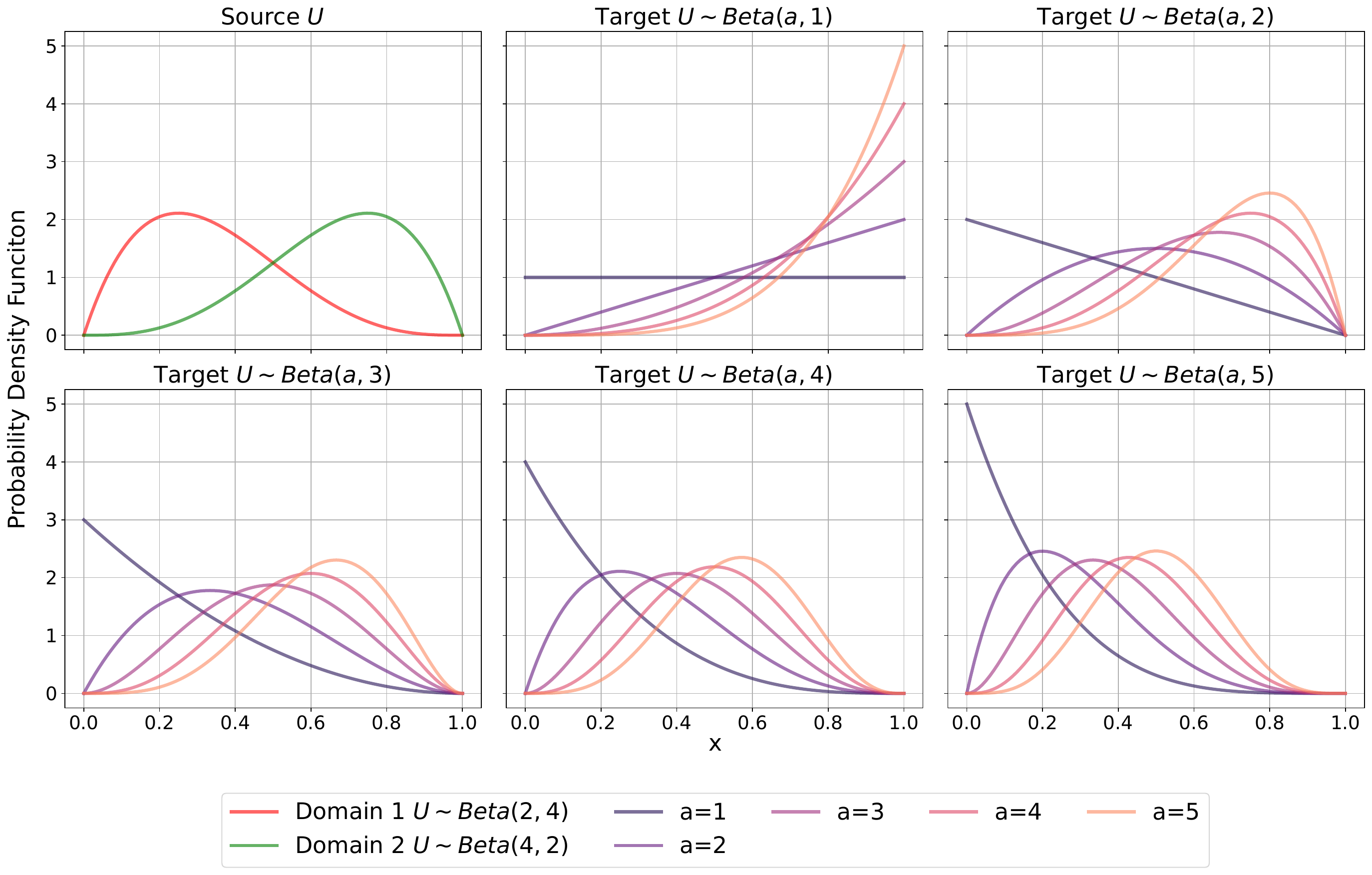}
    \caption{The probablity density function of Beta distributions with different $a,b=1,\ldots,5$. }
    \label{fig:beta_distribution}
\end{figure}

\textbf{Simulated regression task 2.}
\begin{align*}
    U &= Beta(a,b)\\
    X &= \Ncal(0, 1)\\
    Y &= (2U-1)X\\
    W &= \Ncal(-1, 0.01)(1-U) + \Ncal(1, 0.01)U.
\end{align*}
There are two source domains, corresponding to two draws from $P(Z)$ which we write $z_r=(a,b)$. We set $a=2, b=4$ for the first source domain $r=1$, and $a=4,b=2$ for the second source domain $r=2$. The corresponding distributions over $U$ are shown in Figure~\ref{fig:beta_distribution}. Under this setting, we test the target domain with $a,b=1,\ldots, 5$, with distributions shown in Figure~\ref{fig:beta_distribution}. For each domain, we synthesized $2000$ training samples and $1000$ testing samples. We run $10$ replications and the results shown in Figure~\ref{fig:beta_multisource}.

\subsection{Adaptation with Concepts and Proxies}\label{ssec:experiment_cp}

\subsubsection{dSprites Dataset}

\begin{figure}[ht]
    \centering
    \includegraphics[width=0.5\textwidth]{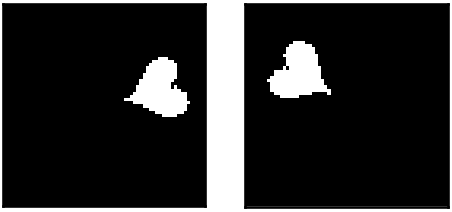}
    \caption{dSprites image with confound (rotation) applied.}
    \label{fig:app_dsprites_img}
\end{figure}

We test the proposed procedure on the dSprites dataset~\citep{dsprites17}, an image dataset described by five latent parameters (shape, scale, rotation, posX, and posY). Motivated by ~\citet{dsprites17}'s experiments, we design a regression task where the dSprites images (64 $\times$ 64 = 4096-dimensional) are $X \in \RR^{64 \times 64}$ and subject to a nonlinear confounder $U \in [0, 2\pi]$ which is a rotation of the image (Figure \ref{fig:app_dsprites_img}). We fix all other latent parameters -- shape is heart, scale is maximized, and all others are set to their 0'th position. $W \in \RR$ and $C \in \RR$ are continuous random variables. The data generation process is defined as follows

\begin{align*}
    U^p &\sim 2\pi\text{Beta}(2, 4),\quad
    U^q \sim \text{Uniform}(a, 2\pi);\\
    X &= \text{Rotate}(\text{image}, U \text{ rads}) + \eta,\quad \eta \sim \Ncal(0, 0.01I_{64}) ;\\
    C &= \Bigg(\frac{0.1\|X^TA\|_2^2 - 5000}{2000}\Bigg)^2 + U + \gamma; \\ 
    A &\sim \text{Uniform}(0, 1),\, A \in \RR^{4096\times 10},\quad \gamma \sim \Ncal(0, 0.5);\\
    Y &= \frac{1}{4}C + \frac{1}{20}\sin(U) + \varepsilon,\quad \varepsilon \sim \Ncal(0, 0.1);\\
    W &= \cos(U) + \nu,\quad \nu \sim \Ncal(0, 0.25).
\end{align*}

When fitting all model, both baselines and the proposed method, we project the images $\RR^{4096}$ to $\RR^{16}$ via Gaussian Random Projection using the {\em scikit-learn} implementation ~\citep{Bingham2001RandomPI, scikit-learn}. Additionally, for the proposed method, we use a Gaussian kernel as the feature map for $X,\, C$. 

We generate $7000$ training samples and $3000$ test samples in our experiments. Then, we use five-fold cross-validation to select hyperparameters for baselines and proposed method for each $a$ ($U^p \sim \text{Uniform}(a,2\pi)$) -- hyperparameters are (i) ridge regression penalty and (ii) Gaussian kernel scaling factor. Once we select a set of hyperparameters for a value of $a$, we perform 10 new random data regenerations to get transfer errors with 95\% confidence intervals (Figure \ref{fig:dsprites_result}).

\subsection{Classification of radiological findings with MIMIC-CXR} \label{appendix:mimic}
We conduct a small-scale experiment with chest X-ray data extracted from the MIMIC-CXR dataset \citep{johnson2019mimic}. 
We consider classification of the absence of a radiological finding in a chest X-ray. 
For this, we use the set of labels extracted by \citet{irvin2019chexpert}.
These labels correspond to 14 categories of radiological findings extracted based on mentions in the associated radiology reports. 
We specifically consider classification of the ``No Finding'' ($Y=1$) label, corresponding to cases where no pathology was identified as positive or uncertain in the radiology report.

To define the dataset, we consider the set of 217,536 chest X-rays with defined Chexpert labels \citep{irvin2019chexpert}, MIMIC-IV entries, and pretrained embeddings \citep{sellergren2022simplified}. We then filter this dataset to the 212,567 examples considered as a part of the ``train'' partition provided by the MIMIC-CXR database \citep{johnson2019mimic}. We then partition the data into training, validation, and testing splits such that 80\%, 10\%, and 10\% of the examples belong to each partition, respectively. For adaptation, we consider BioBERT \citep{lee2020biobert} 768-dimensional embeddings of the radiology reports as concepts $C$ and the patient's age as a proxy variable $W$. For simplicity, we use the patient \texttt{anchor\_age} defined through linkage to the MIMIC-IV database, regardless of the patient's age at the time of the chest X-ray. Similar to the dSprites experiment, we further reduce the dimensionality of $X$ and $C$ to $\mathbb{R}^{64}$ using Gaussian Random Projection fit on the full training partition (170,053 examples).

To define distribution shifts, we adopt a problem formulation similar to that of \citet{makar2022causally}, where patient sex is considered as a possible ``shortcut" in the classification of the absence of a radiological finding. As in \citet{makar2022causally}, we impose distribution shift through structured resampling of the data where $P(U=1) = P(Y = 1 \mid \textrm{Sex}=\textrm{Female}) = P(Y = 0 \mid \textrm{Sex}=\textrm{Male})$. For example, when $P(U=1)=0.1$, the prevalence of $P(Y = 1 \mid \textrm{Sex} = \textrm{Female})=0.1$ and $P(Y = 1 \mid \textrm{Sex} = \textrm{Male})=0.9$. We implement the shift through a weighted sampling procedure that maintains the label shift invariance within patient sex subgroups, i.e., preserves $X \mid Y, A$ under the distribution shift, where $A$ corresponds to patient sex. This procedure further fixes the total proportion of male and female patients in the population at 50\%.
For our experiments, we consider nine domains corresponding to cases where $P(U=1) \in \{0.1, 0.2, \ldots, 0.9\}$.

We perform both concept adaptation and multi-domain adaptation experiments with the MIMIC-CXR data. For the concept adaptation experiment, we perform weighted sampling with replacement of 1,000 examples from each of the training, validation, and testing partitions defined previously, separately for each domain. We fix the source domain to the case where $P(U=1)=0.1$ and then adapt to each of the nine target domains. For the multi-domain adaptation experiment, we randomly sample 500 examples per domain and partition from the sets of 1,000 examples defined for the concept experiment. For this experiment, we consider a case where two source domains corresponding to $P(U=1)=0.1$ and $P(U=1)=0.2$ are available. To match the size of the aggregate source domain data with the size of the target domain, we sample 250 examples per partition for each source domain. We repeat the sampling procedure five times and report the mean $\pm$ standard deviation of performance metrics over the five replicates.

For both experiments, we perform two-fold cross-validation for the kernel length-scale parameters using data from the source domain(s).
Here, we compare to ridge logistic regression models fit in the source and target domains, with the ridge penalty fit with five-fold cross validation. We use \textbf{LR-Target} to refer to logistic regression models fit in a target domain, \textbf{LR-SOURCE} to refer to models fit in a source domain, and \textbf{Cat-LR} to refer to logistic regression models fit with concatenated data from the multiple source domains. We use \textbf{Bridge-SOURCE} to refer to the kernel estimator that leverages the bridge function ($h_0$ or $m_0$ for the concept and multi-domain adaptation settings, respectively) and conditional mean embedding ($\mu_{WC\mid x}$ or $\mu_{W\mid z, x}$) fit on the source domain data. \textbf{Bridge-TARGET} refers to the kernel estimator where both the bridge function and conditional mean embedding are fit on the target domain data. 


\putbib[bu2.bbl]
\end{bibunit}

\end{document}